\documentclass[lettersize,journal]{IEEEtran}
\usepackage{amsmath,amsfonts}
\usepackage{algorithmic}
\usepackage{algorithm}
\usepackage{array}
\usepackage[caption=false,font=normalsize,labelfont=sf,textfont=sf]{subfig}
\usepackage{textcomp}
\usepackage{stfloats}
\usepackage{url}
\usepackage{verbatim}
\usepackage{graphicx}
\usepackage{cite}
\usepackage{orcidlink}
\usepackage{tikz}
\usepackage{stfloats}
\usepackage{booktabs,multirow}
\usepackage{makecell}
\usepackage{color}
\usepackage{enumitem}
\hypersetup{
colorlinks=true,
linkcolor=black,
citecolor=black
}
\newcommand{\re}[1]{{\color{black}{#1}}}

\begin{document}

\title{Mix Q-learning for Lane Changing: A Collaborative Decision-Making Method in Multi-Agent Deep Reinforcement Learning}

\author{Xiaojun Bi, Mingjie He, Yiwen Sun
\hspace{-1.5mm}$^{~\orcidlink{0000-0003-1014-4545}}$
        % <-this % stops a space
% \thanks{Personal use is permitted, but republication/redistribution requires IEEE permission.
\thanks{Copyright (c) 2025 IEEE. Personal use of this material is permitted. However, permission to use this material for any other purposes must be obtained from the IEEE by sending a request to pubs-permissions@ieee.org. 
This work was supported by the key project of the National Natural Science Foundation of China (Grant No.62236011). (Corresponding author: Yiwen Sun.)}
\thanks{Xiaojun Bi and Mingjie He are with the College of Information and Engineering, Minzu University of China, Beijing 100081, China, and Xiaojun Bi is also with the Key Laboratory of Ethnic Language Intelligent Analysis and Security Governance of MOE, Minzu University of China, Beijing, China}
\thanks{Yiwen Sun is with the Institute for Artificial Intelligence, Peking University, Beijing 100871, China, and also with the Beijing Institute for General Artificial Intelligence (BIGAI), Beijing 100080, China (e-mail:sunyiwen@pku.edu.cn).}}

% \markboth{IEEE TRANSACTIONS ON VEHICULAR TECHNOLOGY,~Vol.~14, No.~8, August~2021}%
%\markboth{IEEE TRANSACTIONS ON VEHICULAR TECHNOLOGY}


\maketitle

\begin{abstract}
Lane-changing decisions, which are crucial for autonomous vehicle path planning, face practical challenges due to rule-based constraints and limited data. Deep reinforcement learning has become a major research focus due to its advantages in data acquisition and interpretability. However, current models often overlook collaboration, which affects not only impacts overall traffic efficiency but also hinders the vehicle's own normal driving in the long run. To address the aforementioned issue, this paper proposes a method named Mix Q-learning for Lane Changin\textcolor{black}{g (}MQLC) that integrates a hybrid value Q network, taking into account both collective and individual benefits for the greater good. At the collective level, our method coordinates the individual Q and global Q networks by utilizing global information. This enables agents to effectively balance their individual interests with the collective benefit. At the individual level, we \re{integrate} a deep learning-based intent recognition module into our observation and \re{enhance} the decision network. These changes provide agents with richer decision information and more \textcolor{black}{precise} feature extraction for improved lane-changing decisions. This strategy enables the multi-agent system to learn and formulate optimal decision-making strategies effectively. Our MQLC model, through extensive experimental results, impressively outperforms other state-of-the-art multi-agent decision-making methods, \textcolor{black}{effectively improving the overall efficiency.  The code is available at https:github.com/pku-smart-city/source\_code/tree/main/MQLC}.
\end{abstract}


\section{Introduction}
Cooperative Vehicle Infrastructure Systems (CVIS) involve the dynamic exchange of information between vehicles and road infrastructure using sensing, wireless communication and automated control technologies through vehicle-to-vehicle (V2V) and vehicle-to-infrastructure (V2I) communication. CVIS improves traffic flow, reduces risks and preserves the environment in Intelligent Transport Systems (ITS)\cite{pang2022triboelectric}.

In the field of CVIS, autonomous driving and related applications have the potential to revolutionise transport which could fundamentally change traffic patterns and provide significant convenience\cite{yu2021automated, grigorescu2020survey, zhu2021survey, pei2022fault}. Among the tasks associated with autonomous driving, lane-changing decisions, which \re{involve} a core model that determines whether a vehicle should change lanes based on its state, environment and traffic rules, are prominent. This model also determines the best strategy, aiming to ensure safe and efficient manoeuvres that adapt to current traffic conditions and improve overall performance\cite{wang2022ego}.

The modelling methods for lane-changing decision models can be broadly categorised into three types: Rule-based methods\cite{gipps1986model, kesting2007general, ahmed1996models, sun2012lane, maerivoet2005cellular, schwarting2018planning}, supervised learning-based methods\cite{zhang2022learning, wang2018human, liu2019novel} and deep reinforcement learning-based methods\cite{aradi2020survey}. 
As the fidelity of simulation environments has advanced, deep reinforcement learning-based models hold the promise of completely overcoming their dependence on data and the limitations of mimicking human driving, and ultimately achieving smooth lane changes in various scenarios\cite{li2022decision}. 

\begin{figure}[ht!]
\centering
\subfloat[Illustration of the Importance of Intent Prediction.]{
	\centering
 \includegraphics[scale=0.2]{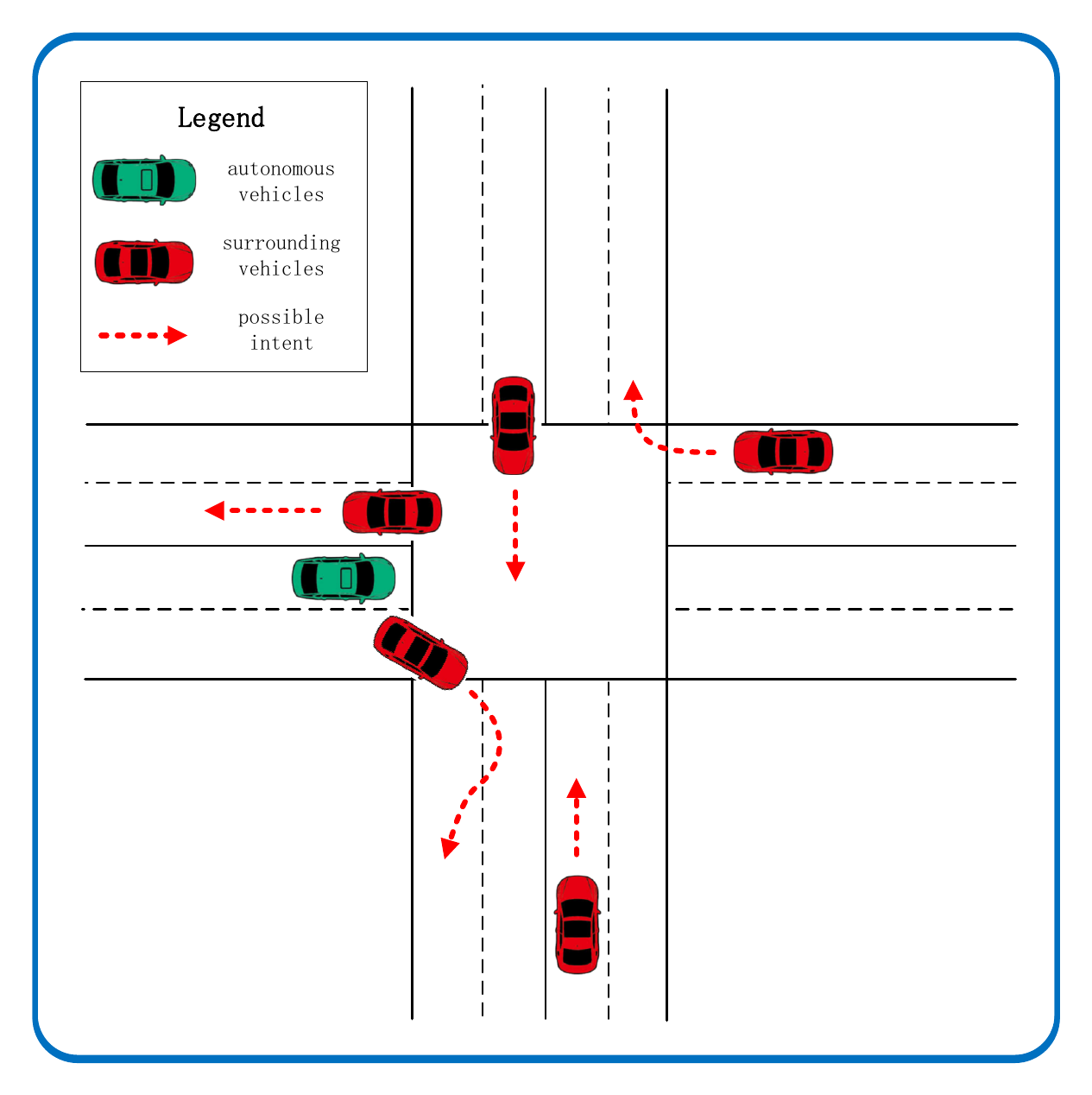}}
\subfloat[Illustration of Importance of Cooperative Driving.]{
\centering
\includegraphics[scale=0.2]{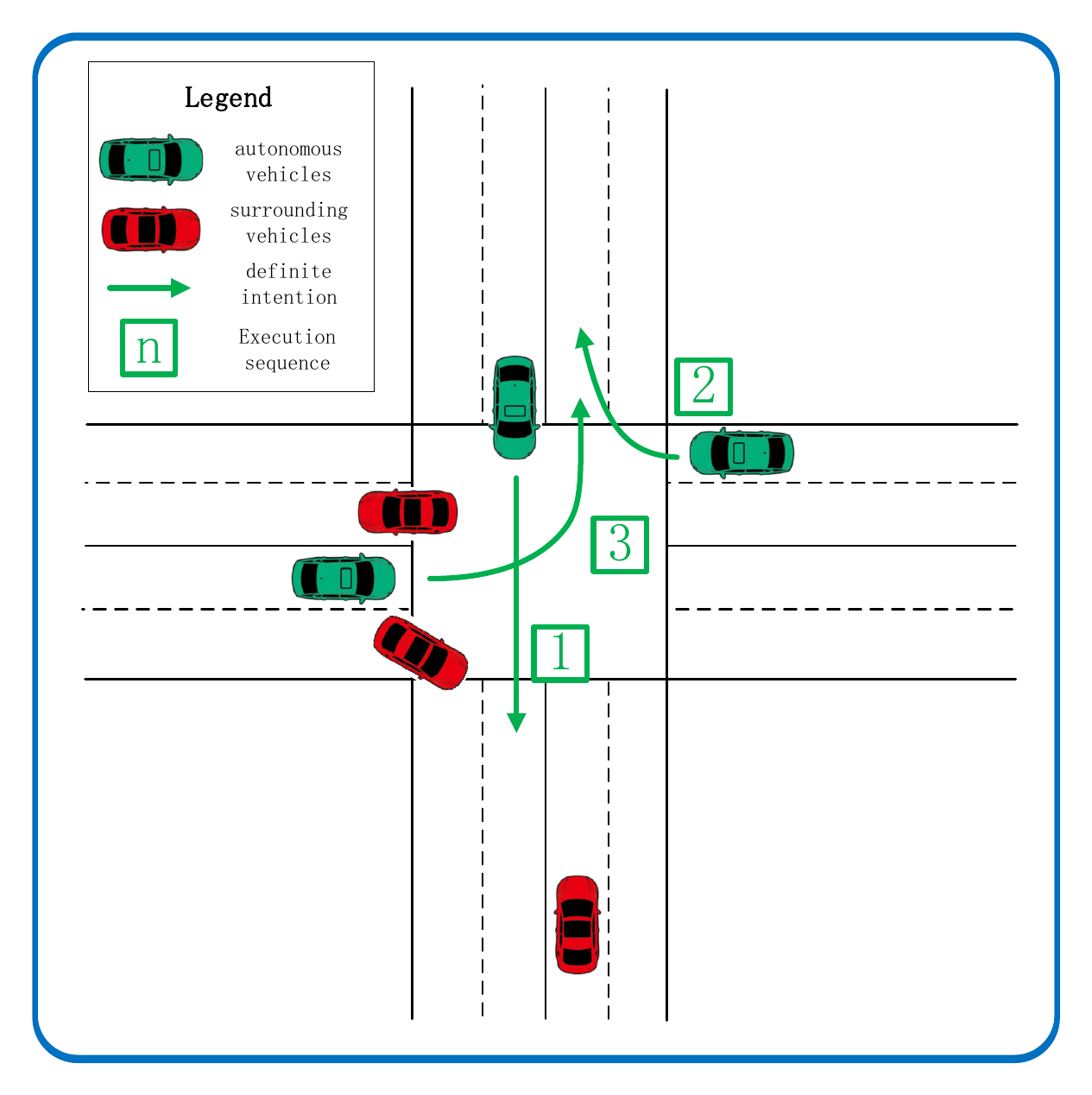}
%\label{Figure 1b}
}
	\caption{The importance of intent prediction and cooperative driving. Figure 1(a) highlights the importance of intent prediction. As autonomous vehicles approach an intersection, the ability to accurately predict the intentions of all nearby vehicles becomes critical. Figure 1(b) highlights the importance of cooperative driving for autonomous vehicles. This scenario shows that when there is a potential conflict between the intentions of the three autonomous vehicles, the unified coordination mechanism enables them to effectively obey traffic rules or resolve the conflict.}
	\label{Figure 1}
 \vspace{-15pt}
\end{figure}

\textcolor{black}{However existing lane-changing decision methods based on deep reinforcement learning have certain defects and limited practical value.}
First, despite the potential for large-scale applications of CVIS\cite{papadimitratos2009vehicular}, most models overlook this crucial aspect and treat lane-changing decisions as independent actions taken by individual agents. Their decisions do not consider collaboration, ignoring the objective interactions between vehicles and the potential for cooperation in the CVIS environment\cite{wang2019lane, alizadeh2019automated}. 
\textcolor{black}{Second, most existing models use generic deep learning decision network architectures, which ignore the specific spatial structure of the interaction between vehicles in the lane-changing decision task\cite{wang2021highway}. }
This generic and underutilised network architecture leads to challenges in understanding the interactions between vehicles and the impact of the decision on the overall traffic situation, ultimately affecting the effectiveness of lane-changing decisions. 
\textcolor{black}{Finally, existing multi-agent lane-changing decision models often employ a centralised training and decentralised execution (CTDE) framework\cite{zhang2022multi, chen2023multi}. 
During execution, the models still follow the single-agent decision-making pattern, considering all surrounding vehicles as part of the environment.} 
Although this decision-making method facilitates learning decisions, it ignores the potential for cooperation between agents in contrast to the centralised training and centralised execution (CTCE) framework. 
A realistic display of the above drawbacks is shown in Figure \ref{Figure 1}. 

In recent years, researchers make significant efforts in multi-agent cooperative decision making, achieving promising results in tasks such as StarCraft\cite{vinyals2017starcraft, usunier2016episodic} and collaborative hunting\cite{tsutsui2022emergence}, and emerge outstanding multi-agent cooperative methods like VDN\cite{sunehag2017value}, $Q_{mix}$\cite{rashid2020monotonic} and DDPG\cite{lillicrap2015continuous}. 
\textcolor{black}{However, due to the complexity, uncertainty, and high sensitivity of lane-changing decision-making scenarios, directly applying the aforementioned methods to lane-changing decisions can lead to mismatches in reward allocation or action space dimensions. }
\vspace{-13pt}

To address the above issues, we propose the Mix Q-learning for Lane Changing (MQLC) model. 
First, we achieve more accurate predictions of neighbouring vehicles' intentions through deep learning-based trajectory prediction, outperforming current techniques. 
In addition, the classical Deep Q-Network (DQN) structure is extended to adapt to the lane-changing decision environment, enabling superior information acquisition and processing at the agent level, thereby facilitating the learning of safe and efficient lane-changing strategies. 
At the decision level, we plan to adopt a collaborative approach between agents in the context of CVIS, using joint decision making with both global and individual coordination. 
Specifically, we adopt the approach proposed by Zhang et al. \cite{zhang2019integrating}, known as the ``QCOMBO" framework, to establish individual and global Q-networks. This framework aims to align the collective Q-decisions of all individuals with the global Q-estimation, thereby incorporating the global context into individual decision-making processes. However, in this framework, individual Q networks were still solely responsible for decision making without taking into account collaborative considerations. To address this limitation and to account for the collaborative nature of lane-changing decisions, we analyse the characteristics of collaboration in lane-changing decisions and introduce the paradigm of individual Q-network action preselection followed by global Q-network arbitration. \textcolor{black}{This paradigm ensures that all agents make coordinated lane-changing decisions while maximizing overall utility and preserving individual gains. }
By enhancing both decision information analysis and decision formulation aspects, our method aims to achieve rational individual control and improve overall traffic efficiency. 

In general, the contributions of this paper are as follows:

1. \textcolor{black}{To the best of our knowledge, we propose the first collaborative decision-making framework for multi-agent vehicles in the field of lane-changing decision-making. Through the cooperation of individuals and the global network,} MQLC achieves the maximisation of overall benefits while ensuring the basic lane-changing gains of each agent. 

2. \textcolor{black}{We have achieved intention prediction based on deep neural networks that can better understand the interactions between vehicles, thereby enriching the decision process with valuable information.} In addition, we improve the structure of the reinforcement learning network by combining the advantages of multiple neural networks to make it more suitable for lane-changing decision tasks. This ensures that the decision network can make more intelligent lane change decisions \textcolor{black}{and improve overall traffic efficiency.}

3. We conduct experiments on MQLC in a simulation environment. The results show that our model consistently produces reasonable and safe lane-changing decisions under different traffic conditions. It outperforms other state-of-the-art lane-changing decision models and multi-agent cooperative models in various evaluation metrics.

The remainder of this paper is structured as follows. Section \uppercase\expandafter{\romannumeral2} introduces the related work. Section \uppercase\expandafter{\romannumeral3} presents the formal problem definition of the lane-changing task and the specific design of the MQLC model. Section \uppercase\expandafter{\romannumeral4} presents the experimental results and discussions. Finally, we provide a summary of our work in section \uppercase\expandafter{\romannumeral5}.

\section{Related work}

According to the methods used for model formulation, current lane-change decision models can be broadly classified into two types: rule-based models and learning-based models. 
Among the learning-based models, they can be further divided based on whether they use real human driver vehicle motion data, leading to two subcategories: supervised learning-based models and deep reinforcement learning-based models. 
The following section introduces the dynamics of lane-changing decision models from these three perspectives.
\subsection{Rule-based models}
Rule-based models aim to use predefined sets of rules and heuristics to guide lane-changing decisions. Their creation and design \re{are} based on expert knowledge.
Rule-based models have attracted considerable attention from researchers due to their ease of application and the high interpretability of their decision processes \cite{guo2022lane}. Within this category, classic models include the City Driving Decision Structure introduced by Gipps et al.\cite{gipps1986model} and the generalised vehicle-following model for urban traffic further improved by Kesting et al.\cite{kesting2007general}. 
%Other modelling methods include the use of Traffic Cellular Automaton (TCA) models, as demonstrated by Maerivoet et al.\cite{maerivoet2005cellular}, and driver behaviour classification methods, as explored by Sun et al.\cite{sun2012lane}. These modelling methods help to better simulate lane changing behaviour in urban traffic environments.

However, rule-based models rely on expert knowledge and suffer from limited flexibility and generalisation, making them unsuitable for complex and dynamic road scenarios.% As a result, researchers have shifted their focus to learning-based models.

\subsection{Supervised learning-based models.}
Supervised learning-based models typically use deep neural networks to regress human vehicle trajectory data to learn lane-changing decisions similar to human drivers.
In recent years, supervised learning methods have emerged to address lane-changing problems, such as the use of Long Short-Term Memory (LSTM) neural networks combined with Conditional Random Fields (CRF) to learn human-like manoeuvring decisions\cite{wang2018human}, and the use of Bayesian parameter optimisation and Support Vector Machine (SVM) techniques\cite{liu2019novel}.
%To enrich limited decision information, Zhang et al.\cite{zhang2022learning} integrated human factors represented by driving style into the study of autonomous lane changing (DLC) to achieve more rational lane changing decisions. 

While supervised learning methods have achieved some success, they face significant challenges related to the collection and processing of human driver trajectory data%, hindering the development of such methods
. In addition, human lane-changing decisions are based on limited observations, and personality traits of human drivers can negatively influence these decisions. As a result, researchers have shifted their focus to reinforcement learning-based lane-changing decision methods.

\subsection{Deep reinforcement learning-based models.}
The deep reinforcement learning-based decision models learn rich decision knowledge through interactive trial-and-error experience in a simulated environment. 
This approach avoids learning overly aggressive or conservative negative decision behaviours exhibited by human drivers, and mitigates the dependence on human driver trajectory data, which is a limitation of supervised learning-based models. 
As a result, deep reinforcement learning-based models emerge as the primary research direction for lane-changing decision models.

Based on the innovations and contributions of these methods, existing DRL-based lane-changing models can generally be classified into three types.
The first category focuses on improving performance by enhancing the input of the model, such as using attention mechanisms\cite{chen2019attention, wang2021highway} or adding Bird's Eye View (BEV) images with vector\cite{wang2021highway} input to boost information extraction capabilities.
%The first category focuses on the input of the model, aiming to improve performance by designing the observation space or incorporating specific network structures at the input end. 
%For example, introducing an attention mechanism at the input end\cite{chen2019attention, wang2021highway} or adding a Bird's Eye View (BEV) image with vector input\cite{wang2021highway} to improve the model's information extraction capabilities.
% More recently, there has been a method that uses observed adversarial research to improve the robustness of the lane-changing decision model\cite{he2022robust}.
The second category improves performance by designing better reward functions. Recent approaches include incorporating global traffic flow\cite{wang2021harmonious} or risk terms\cite{li2022decision} into the reward function to make lane-changing models more harmonious and safer.
%The second category emphasises the reward function, using cleverly designed reward functions to improve performance.
%Recently, there have been innovative ways to introduce a global traffic flow\cite{wang2021harmonious} or risk term\cite{li2022decision} into the reward function to make the lane-changing model more harmonious and safer when changing lanes. 
% Yuan et al.\cite{yuan2019multi} decomposed the traditional single reward function into three sub-rewards, simplifying the learning difficulty of the lane-changing task.
The third category emphasizes the algorithm and framework, adding safety layers like safety verification\cite{mirchevska2018high} or safety supervisors\cite{chen2020autonomous} to ensure smooth operation in real-world situations.
%The third category focuses on the algorithm and framework of the lane-changing model. 
% Xu et al.\cite{xu2018reinforcement} proposed a multi-objective approximate policy iteration algorithm to improve the safety, efficiency, and smoothness of lane-changing behaviour for autonomous vehicles. 
%To add an extra layer of safety, the researchers incorporate structures such as safety verification\cite{mirchevska2018high} or safety supervisors\cite{chen2020autonomous} into the algorithmic framework to ensure that the model runs smoothly in real-world situations.

MQLC is \re{deep} a reinforcement learning-based lane-changing decision method that simultaneously improves both input and framework aspects. %Compared to other reinforcement learning methods, 
It achieves explicit coordination between lane-changing agents through additional inputs and a collaborative decision-making process.

\section{Methodology}

\subsection{Problem formulation}
\label{sec:3.1}
\re{
Based on the previous statement, we address the lane-changing decision problem in a multi-agent scenario. Our goal is to improve information acquisition, decision networks, and decision architecture to enable agents to achieve better operating conditions through more rational lane-changing strategies. We formulate the multi-agent lane-changing decision problem as a Partially Observable Markov Decision Process (POMDP) follow\cite{zhang2022multi}. In addition to the components of a regular MDP, such as the state space $ \mathcal{S} $, the action space $\mathcal {A}$, the reward function $ \mathcal{R} $, and the state transition model $ \mathcal{P} $, POMDP introduces an observation space $ \Omega $ ($ o_t \in \Omega $) and an observation probability distribution $\mathcal{O}\left(o_{t} \sim \mathcal{O}\left(s_{t}\right)\right)$ representing the observations in state $ s_t $. Based on the POMDP model, each vehicle independently samples observations from the state and autonomously executes actions, receiving rewards provided by the environment, to progressively discover the strategy that can achieve the highest future rewards.

Observation space: The observation space is the collection of observations on which the vehicle’s decision making is based. As mentioned earlier, the observation at time t for
our vehicle consists of two parts: kinematic observations $ \Omega_k $ and driving intentions $ \Omega_I $.

Kinematic observation: $ \Omega_k $ represents the kinematic state information of the ego vehicle and its surrounding vehicles, denoted as $ K_{i}^t $ with a size of $ V $×$ F $. Here, $ V $ represents the number of observable vehicles, including the ego vehicle itself. $ F $ corresponds to the set of vehicle features, represented as presence, $ x $, $ y $, $ v_x $, and $ v_y $, where the five values indicate whether the vehicle is alive, its position and velocity.

Driving intentions: $ \Omega_I $ aims to enable the ego vehicle to consider the potential future driving actions of surrounding vehicles when making decisions. We propose a novel approach for intention prediction. We encode the scene from the previous time step $ T_P $ and feed it into a neural network to predict the coordinates of the surrounding vehicles at a future time step $T_F$ . Compared to polynomial-based methods\cite{zhang2022multi}, deep learning based trajectory prediction offers stronger capabilities in nonlinear modelling, adaptability and generalisation. The specific network architecture is shown in Table I of the experimental section. The reason for choosing this architecture is that the vehicle coordinate trajectory is input into the network in the form of points. GCN can better extract the topological structure features of the vehicle, while GRU is used for time series trajectory prediction.

Action space: for vehicle agent i at time t, the action $ a $ it takes is sampled from its advantage estimation over all feasible actions in the discrete action space. And the action space $\mathcal {A}$ consists of five discrete actions: 

\begin{equation}
    a_t \in \mathcal {A} = \{lane_{left}, idle, lane_{right}, faster, slower\}.
\end{equation}

The joint action space is the set of all possible actions for all vehicles, represented in matrix form as $A_{joint} = A_1 \times A_2 \times \cdots \times A_n$, where n is the number of agent vehicles.

Reward function\cite{highway-env}: The reward is a signal provided to the agent after it takes an action. This signal reflects changes in the environment resulting from the state transition model. This paper hopes to verify the effectiveness of the proposed collaborative method in lane-changing decision-making, so the default reward items and functions in the virtual environment\cite{highway-env} are selected.
\begin{equation}
    R = \omega_{1}R_1 + \omega_{2}R_2 + \omega_{3}R_3 ,
\end{equation}
where $ R_1 $, $ R_2 $, and $ R_3 $ represent the safety reward term, unnecessary lane change penalty term, and high-speed reward, respectively. $ \omega_1 $, $ \omega_2 $, and $ \omega_3 $ are the respective coefficients for these three reward terms.
}

\subsection{Deep Reinforcement Learning Algorithm}
\begin{figure}[ht!]
	\centering
	\includegraphics[scale=0.25]{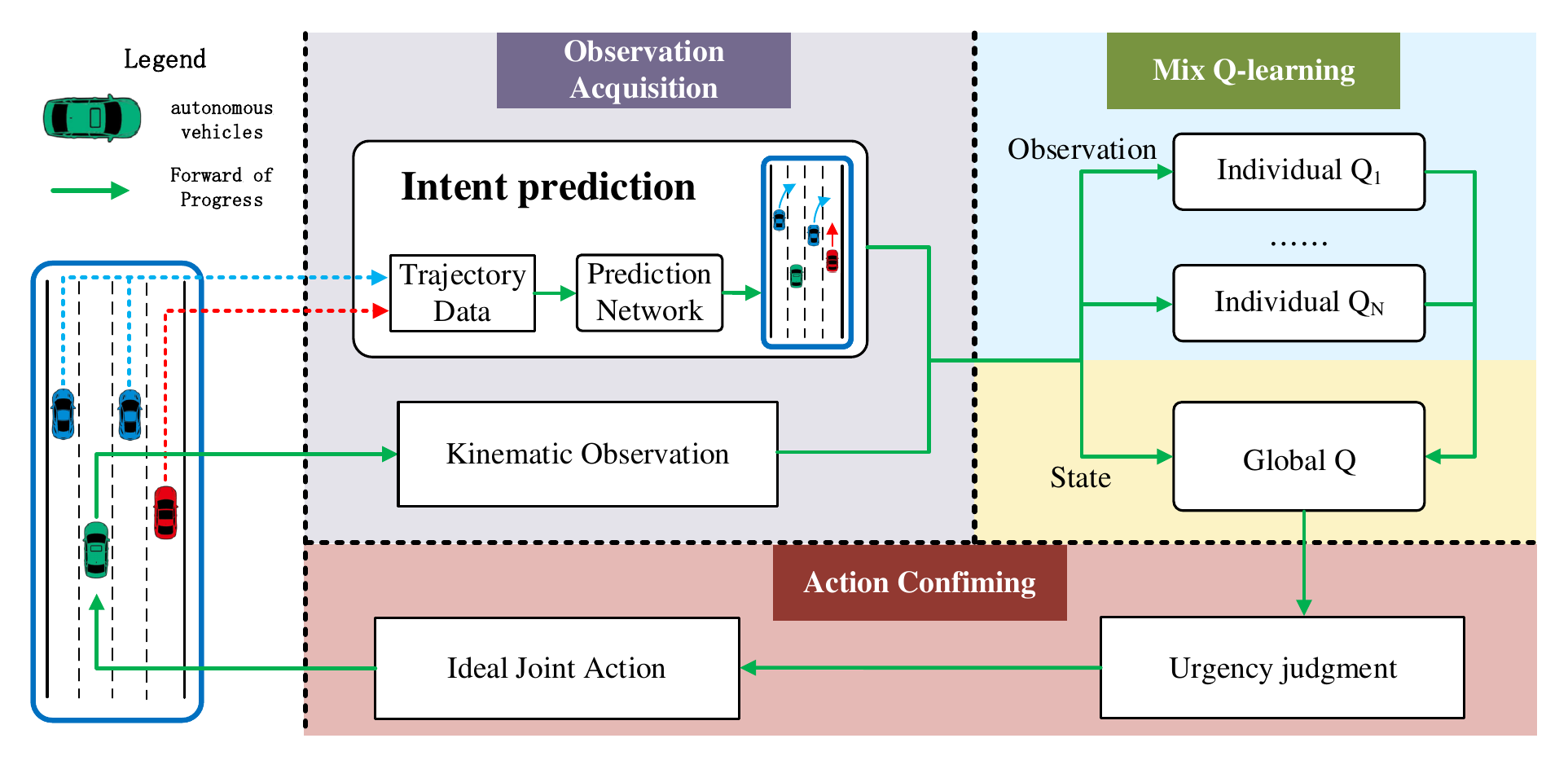}
	\caption{The overall framework of MQLC. The decision making process of the intelligent agent can be described as follows: first, MQLC estimates the intentions of the surrounding vehicles based on their trajectories. Next, at the individual level, the agent uses an innovative network structure to optimise information embedding and estimate feasible decisions. Finally, at the global level, an advantage estimation is performed for the joint actions that all intelligent agents can take.% This estimation guides the intelligent agents in achieving collaborative lane-changing decisions.
    }
	\label{Figure 2}
\end{figure}

In the context of the lane-changing decision task, reinforcement learning algorithms seek to balance adaptability and exploratory behaviour.
To achieve such a lane-changing decision algorithm, this paper proposes a novel collaborative lane-changing algorithm and makes modifications to the network architecture to make it more suitable for the lane-changing decision task.
The overall framework of MQLC is shown in Figure \ref{Figure 2}.
A detailed explanation is given below. 

\subsubsection{collaborative lane-changing algorithm framework}
Although many reinforcement learning methods have been applied to lane-changing decision tasks and achieved considerable success, their improvements usually focus on reducing estimation errors, improving stability, convergence, etc., without considering multi-agent collaboration.

In order to achieve true collaboration in lane-changing decisions, we \re{reference} QCOMBO\cite{zhang2019integrating} algorithm proposed for signal light optimisation when designing the algorithm framework. We then made modifications to adapt it to the characteristics of the lane-changing task.
QCOMBO is a value-based reinforcement learning network. Compared to DQN, it combines centralised learning and independent learning, which is manifested in its individual Q and global Q.
The optimisation objective of \re{QCOMBO} consists of three parts. 
In addition to the losses from the two reward-feedback networks, an additional regulariser is introduced to minimise the difference between the weighted sum of the global Q and the individual Q. 
The specific design of the losses is as follows:

\textit{(a) Individual part:} 
The training goal for the individual part is to gradually make the individual Q network's advantage estimates for each action in the observation approach the true feedback reward, based on the agent's interactions in the environment. 
To improve the training efficiency and stability, a strategy of using local reward\cite{sunehag2017value} and parameter sharing\cite{gupta2017cooperative, chu2017parameter} is adopted in this part.
%In terms of lane-changing decisions, agents share the same observation and action space. By using parameter sharing in an execution, the environment can obtain multiple data sets. 
%This allows for a richer data set to train the Q-network, thereby improving training efficiency. 

For the individual Q-network of agent $i$ parameterised by $\theta^{i}$, minimise the following loss
\begin{equation}\label{6}
\mathcal{L}_{ind}(\theta^{i}) = \mathbb{E}_{\left(o_t^{i}, a_t^{i}, r_t^{i}, o_{t+1}^{i}\right) \sim D_{ind}}\left[\left(y_t^{i}-Q_{i}\left(o_t^{i}, a_t^{i} ; \theta^{i}\right)\right)^{2}\right], \\
\end{equation}
\begin{equation}\label{7}
y_t^{i} = r_t^{i}+\gamma \max _{a_{t+1}^{i}} \hat{Q}_{i}\left(o_{t+1}^{i}, a_{t+1}^{i} ; \theta^{i}\right),   
\end{equation}
where $D_{ind}$ is the individual experience replay buffer, $o_i$ is the observation, $a_i$ is the action, $r_i$ is the reward, $o_{t+1}^{i}$ is the next observation, $\gamma$ is the discount factor, $\hat{Q}_i$ is the target Q-network, and ${Q}_{i}$ is the individual Q-network. 

\textit{(b) Global part:} 
The purpose of the global part is to train a global Q network that can provide estimates of the benefits of joint actions based on global information and come close to the total reward feedback.
%Due to the necessity of selecting the best action out of n possible joint actions during the learning process of Q networks, the design of the global Q is limited in scalability in many multi-agent tasks. 
%To address this issue, QCOMBO proposes the idea of training the global Q using local utility functions. 
The global Q treats the action taken by each agent's individual Q when it is optimal as the joint action and only estimates the utility of that action.
Specifically, the joint action is chosen by $\mathbf{a} \sim \boldsymbol{\pi}(\mathbf{a} \mid s)=\left\{\operatorname{argmax}_{a^{i}} Q^{n}\left(o^{i}, a^{i}\right)\right\}_{i=1}^{N}$, and its value is defined by $R_t^g:=R_t^g(s,\mathbf{a})$. Based on the above, the global Q can be written as 
\begin{equation}\label{8}
Q^{\boldsymbol{\pi}}(s, \mathbf{a}):=\mathbb{E}_{\boldsymbol{\pi}}\left[\sum_{t=0}^{\infty} \gamma^{t} R^{g} \mid s_{0}=s, \mathbf{a}_{0}=\mathbf{a}\right]. 
\end{equation}
For the global Q-network parameterised with $\omega$, minimise the following loss
\begin{equation}\label{9}
\mathcal{L}_{glo}(\omega ) = \mathbb{E}_{\left(s_t,\mathbf{a}_t\right) \sim D_{glo}}\left[\left(y_t-Q_{\omega}^{\pi } \left(s_t,\mathbf{a}_t\right)\right)^{2}\right], 
\end{equation}
\begin{equation}\label{10}
y_{t}=R_{t}^{g}+\left.\gamma Q_{\hat{\omega}}^{\boldsymbol{\pi}}\left(s_{t+1}, \mathbf{a}_{t+1}\right)\right|_{a_{t+1}^{i}=\operatorname{argmax}_{a^{i}} Q_{\hat{\theta}}^{i}\left(o_{t+1}^{i}, a^{i}\right),} 
\end{equation}
where $\gamma$ is the discount factor, $D_{glo}$ is the global experience replay buffer, $s_t$ is the state, $\mathbf{a}_t$ is the joint action, $\hat{\omega}$ are the target network parameters, $s_{t+1}$ is the next state and $\mathbf{a}_{t+1}$ is the next joint action. 

\textit{(c) Combined objective:} 
\label{chap:3.2.1}
Based on the previous concept, the global Q evaluates the actions sampled from the value judgments of the individual Qs for each agent based on global information, but its actual decision making is entirely dependent on the individual part. 
However, if each individual Q takes its optimal action, it may not lead to the best overall performance \cite{sunehag2017value}.
To enable individual Qs to take the global context into account when making decisions, and to achieve effective coordination with other agents to improve overall performance, QCOMBO introduces a consistency regularisation loss between the global $Q_{\boldsymbol{\omega }}$ and the individual $Q_{\boldsymbol{\theta}}$:
\begin{small}
\begin{equation}\label{11}
\mathcal{L}_{reg}:=\mathbb{E}_{\pi}(s_t,\mathbf{a}_t,o^{i}_t, a^{i}_t)\sim D\left[\left(Q_{w}^{\pi}(s_t, \mathbf{a}_t)-\sum_{i=1}^{N} k^{i} Q_{\theta}^{i}\left(o^{i}, a^{i}\right)\right)^{2}\right]. 
\end{equation}
\end{small}
\textcolor{black}{Through this consistency regularization loss, the weighted sum of the action advantage estimates of the individual Q-networks of all agents gradually approximates the true global reward based on global information. 
This enables agents to implicitly consider the concept of the estimated sum approaching the global reward in their decision-making process. 
In other words, the individual decisions of the agents are endowed with a sense of global coordination, rather than merely focusing on individual benefits.}

The total loss can be represented as follows
\begin{equation}\label{12}
\mathcal{L}_{tot}(\omega,\theta) = \mathcal{L}_{glo}(\omega) + \mathcal{L}_{ind}(\theta)+ \lambda\mathcal{L}_{reg},  
\end{equation}
where $\lambda$ is the weight of the regularisation term, and its value determines the degree to which cooperation is taken into account. 
A higher value of $\lambda$ indicates a stronger emphasis on collaboration, causing the weighted sum of individual Q-estimates to be closer to the global Q-estimate. 

Despite QCOMBO's integration of global information and coordination between two hierarchical networks to account for collaboration in multi-agent decision making, it still faces two problems. 
First, QCOMBO's decisions rely entirely on individual Q-values, which in turn depend on the observations of each agent. 
If there is no communication between agents, the lack of complete information in the input may prevent agents from determining specific collaboration methods and hinder true cooperation between agents.
In addition, the design of global Q  estimates the global advantage by taking the optimal action from individual Q values, rather than searching across all joint actions. 
This design limits the global Q to extracting environmental information only from the state, without considering the advantages of joint actions among the available actions. This limitation ultimately affects the overall decision performance.

To address these limitations, we introduce an innovative decision-making process in the MQLC framework, where coordination is achieved through the collaborative efforts of two-level Q-networks. 
%Specifically, in a single decision instance within the MQLC framework, each agent first uses its individual Q to evaluate the benefits of different actions based on its own observations. 
To ensure the generalisability and reliability of MQLC, we introduce an additional urgency concept, i.e. the decision priority, by considering the surrounding traffic conditions. 
The decision priority is defined by the following formula,
\begin{equation}\label{13}
priority_i=\left\{\begin{array}{ll}
high, & { urgency_i>\varepsilon } \\
low, & { urgency_i\le \varepsilon },
\end{array}\right.
\end{equation}
\begin{equation}\label{14}
urgency_i = m \_velocity_i + t\_density_i + \alpha s\_variance_i ,
\end{equation}
where $priority_i$ is the decision priority of the agent $i$, $urgency_i$ is the quantified level of congestion, $m \_ velocity$ is the average speed of the agent and the surrounding vehicles, $t\_density$ is the traffic density in the environment surrounding the agent, while \re{$s\_variance$} is the variance in speeds between the agent and the surrounding vehicles. $\alpha$ is the weight of the speed variance term. 
In this way, MQLC can adapt to different complex traffic scenarios by simply adjusting the urgency parameters, without having to repeat complex reinforcement learning training. At the same time, the priority concept, which is independent of the Q-network, can ensure the basic interests of the agent, rather than completely obeying the whole for the highest global reward.

The definition of the above decision priority is based on the following assumptions: first, as the traffic situation becomes more complex, the agent needs to make lane-changing decisions faster to avoid risks or obtain better driving conditions. 
The assessment of the traffic situation is based on the following factors: the higher the average speed of the agent and the surrounding vehicles, the shorter the reaction time for potential dangers ahead, given a fixed braking capability, indicating a more urgent traffic situation; the denser the surrounding vehicles, the more complex the situation, requiring more cautious decisions; the higher the speed variance of the agent and the surrounding vehicles, the more abnormal driving situations occur in the traffic, reflecting a more complex environment. Among these factors, the speed variance term can best reflect the complexity of the traffic conditions and is therefore given a weight of $\alpha>1$ to emphasise its importance.

Then, based on the decision priority, the agent samples the optimal action if it has a high priority. 
For agents with lower priority, the top n optimal actions are sampled and sent to the global Q. 
The global Q evaluates the benefits of these feasible joint actions derived from the individual Q sampling based on the state s, and selects the group of joint actions with the highest expected benefit for execution. 
The execution and update processes of MQLC can be found in Algorithm \ref{MQLC execution} and Algorithm \ref{MQLC upgrade} respectively. 

\re{
In the MQLC method proposed in this article, there is no correlation between individual Q (-network) and global Q (-network) at the neural network architecture level, that is, there is no interpenetration or stacking between the layers of the two networks. The correlation between these two neural networks lies in the fact that the output of individual Q is a part of the input of global Q, which affects the output of global Q. Specifically, each agent determines the optimal and suboptimal actions that can be taken based on individual Q. The global Q selects one of the optimal and suboptimal actions that each agent can take based on the global situation, ultimately selecting the joint action that maximizes global expected benefits.
}
\begin{figure}[t!]
	\centering
	\includegraphics[scale=0.5]{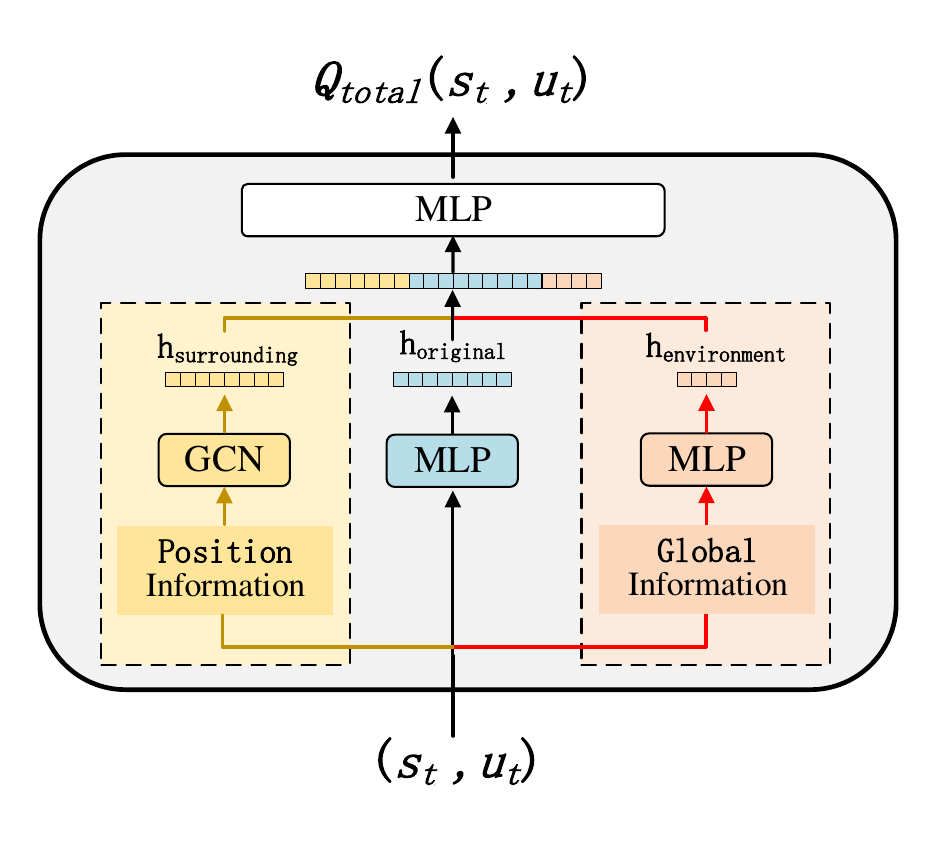}
	\caption{The architecture of the MQLC decision network. The network architecture of the global Q function is presented. In this case, when the input is $(o_t^i, a_t^i)$, the network transforms into an individual Q network that estimates discrete actions based on observations. For any given network, it additionally extracts vehicle position information and global traffic flow information from the input. This information is encoded using GCN and MLP respectively. The final output is an estimate of the benefits of the actions.}
	\label{Figure 3}
\end{figure}

\begin{algorithm}[ht!]
\caption{The execution process of Mix Q-learning for Lane Changing}
\label{MQLC execution}
\begin{algorithmic}[1]
\renewcommand{\algorithmicrequire}{\textbf{Input:}}
\REQUIRE individual value network $Q_{\boldsymbol{\theta}}$, global value network $Q_{\boldsymbol{\omega }}$, individual experience replay buffers $D_{ind}$ and global experience replay buffers $D_{glo}$
\renewcommand{\algorithmicensure}{\textbf{Output:}}
\ENSURE an estimate for discrete actions and an estimate for joint actions

set target networks' parameters the same as the original network:$$\bar Q_{\boldsymbol{\theta}} \leftarrow Q_{\boldsymbol{\theta}}, \bar Q_{\boldsymbol{\omega }} \leftarrow Q_{\boldsymbol{\omega }}$$

\FOR{$iteration=1$ \TO $M$}
    \STATE $//Collect \ training \ data$
    \FOR{$Vehicle \ i=1$ \TO $N$}
        \STATE Compute the expected return of actions in action space.
        $$Q_{\boldsymbol{\theta}}\leftarrow\mathbb{E}_{\boldsymbol{\theta}}\left[\sum_{t=0}^{\infty} \gamma^{t} r(o_t^i,a_t)\right]$$
        %\IF{$arr[j] > arr[j+1]$}
        %    \STATE Swap $arr[j]$ and $arr[j+1]$
        %\ENDIF
        \STATE Calculate the priority of decision-making according to Eq.\eqref{13} and Eq.\eqref{14}.
        \IF {High decision priority}
            \STATE Sampling the optimal action $a_t^i$
        \ENDIF
        \IF{Low decision priority}
            \STATE Sampling the top n optimal actions $a_t^i$
        \ENDIF
        get joint action $\mathbf{a}_t$ from the combination of $a_t^i$
    \ENDFOR
    \STATE Compute the expected advantage of feasible joint action
    $$Q_{\boldsymbol{\omega }}\leftarrow\mathbb{E}_{\boldsymbol{\pi}}\left[\sum_{t=0}^{\infty} \gamma^{t} r(s_t,  \mathbf{a} )\right]$$
    \STATE sample the best joint action $\mathbf{a}_t$
    \STATE Execute and collect transitions $(o_t^i,a_t^i,r_t^i,o_{t+1}^i)$ and $(s_t, \mathbf{a}_t, \mathbf{r}_t, s_{t+1})$.
    \STATE Store transitions in  $D_{ind}$ and  $D_{glo}$.
\ENDFOR
\end{algorithmic} 
\end{algorithm}

\begin{algorithm}[ht!]
\caption{The update process of Mix Q-learning for Lane Changing}
\label{MQLC upgrade}
\begin{algorithmic}[1]
\renewcommand{\algorithmicrequire}{\textbf{Input:}}
\REQUIRE individual value network $Q_{\boldsymbol{\theta}}$, global value network $Q_{\boldsymbol{\omega }}$, individual experience replay buffers $D_{ind}$ and global experience replay buffers $D_{glo}$
\renewcommand{\algorithmicensure}{\textbf{Output:}}
\ENSURE an estimate for discrete actions and an estimate for joint actions
    \STATE Update the individual and global value networks.
    \FOR{$iteration=1$ \TO $M$}
        \FOR{Learning step=1 \TO $K$}
            \STATE sample a batch of samples from replay buffers
            $$B_{ind}=\{(o,a,r,o^{'})\} \; from \; D_{ind}$$ 
            $$B_{glo}=\{(s,\mathbf{a},\mathbf{r},s^{'})\} \; from \; D_{glo}$$ 
            \STATE Calculate the targets by Eq.\eqref{7} and Eq.\eqref{10}
            \STATE Update the value networks
            $$\theta \leftarrow \theta-\lambda _Q\bigtriangledown _\theta J(\theta)$$
            $$\theta \leftarrow \omega-\lambda _Q\bigtriangledown _\omega J(\omega)$$
            \STATE Update the target networks
            $$\bar Q_{\boldsymbol{\theta}} \leftarrow Q_{\boldsymbol{\theta}}, \bar Q_{\boldsymbol{\omega }} \leftarrow Q_{\boldsymbol{\omega }}$$
        \ENDFOR
    \ENDFOR
\end{algorithmic}
\end{algorithm}

\subsubsection{The specific design of the value network}
The two value networks we propose are both based on deep neural networks. 
The individual Q network takes the observation $o$ as input and outputs payoff estimates for individual actions. 
On the other hand, the global Q network takes the state $s$ as network input and outputs benefit estimates for joint actions that can be taken.
Given the matrix input form, both networks use Multi-Layer Perceptron (MLP) as their main components. 
MLP uses its strong feature extraction and generalisation capabilities to fulfil the specific functions of value networks.
Although MLP is widely used in decision learning, it has inadequate information extraction capability limitations, which leads to less efficient decision making. 
To overcome these limitations, MQLC uses a combination of several network architectures in the decision making process. The MQLC decision network architecture is shown in Figure \ref{Figure 3}. 

For both the input observation $o$ in the individual Q-network and the input $s$ in the global Q-network, MQLC follows the same processing procedure. 
First, taking advantage of the excellent capabilities of Graph Convolutional Neural Networks (GCN) to handle topological graph structures, the vehicle coordinate position information in $o$ or $s$ is extracted and passed through two layers of GCN to obtain hidden vectors $h_{sur}$ representing the spatial structure of the surrounding vehicles.
Next, the average speed and traffic density of the surrounding vehicles, derived from the vehicle speed information, are transformed by an MLP into hidden vectors $h_{env}$ representing the environmental influences. 
After obtaining $h_{sur}$, $h_{env}$ and $h_{ori}$ from original observation, these three hidden representations are combined. 
The combined output is then passed through a ReLU activation layer and further processed by a fully connected layer to estimate the advantage for discrete actions or joint actions.

\section{Performance evaluation}
In this section, we perform experiments and analyses of the proposed method. 
Subsection \ref{sec:4.1} introduces the simulation environment and the parameter settings in the experiments. Subsection \ref{sec:4.2} presents the result of the comparative experiments. Subsection \ref{sec:4.3} discusses the parameter settings of the training state. Section \ref{sec:4.4} performs ablation experiments to explore the role of different components of the model. Subsection \ref{sec:4.5} conducts qualitative experiments to investigate the effect of cooperative strategy and emergency situation settings on decision making. 

\subsection{Simulation environment and parameters settings}
\label{sec:4.1}
\begin{figure}[ht!]
	\centering
	\includegraphics[scale=0.45]{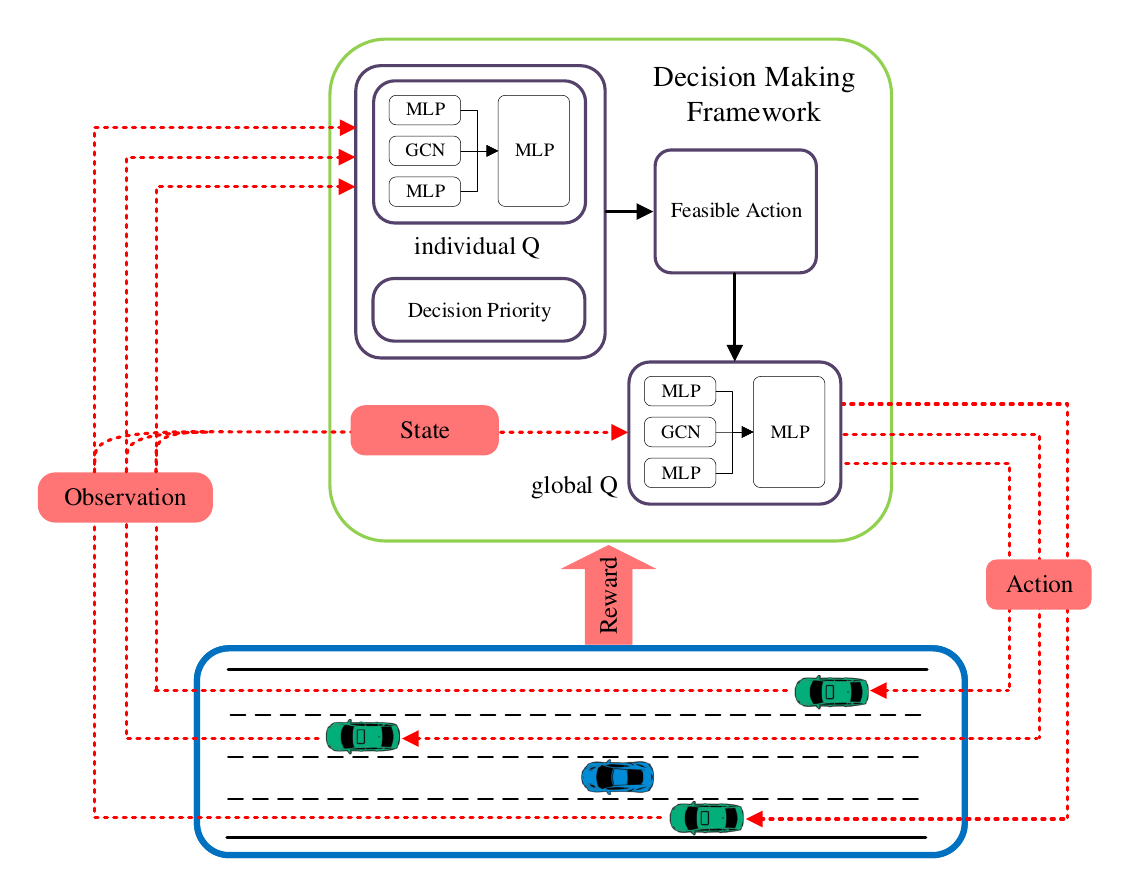}
	\caption{An illustration of the execution process. \textcolor{black}{This figure is a graphical explanation of Algorithm 1.} During the execution phase, the agent's observations are fed into the decision framework. First, each intelligent agent evaluates the benefit values of different actions and prioritises its own decisions based on the observations. Then, a few feasible actions are selected probabilistically. The global Q-function evaluates the benefits of joint actions based on the state formed by aggregating all observations, and finally selects the optimal joint actions for execution by each individual agent.}
	\label{Figure 4}
\end{figure}

To evaluate our proposed lane-changing decision model, we used OpenAI's gym-based highway-env simulator\cite{highway-env} to construct a multi-lane highway traffic environment. 
The simulator provides the flexibility to adapt highway scenarios as needed and includes settings to access global information and make unified decisions, in line with the focus of our study on individual-global collaborative decision making. 
We deployed the MQLC model within the simulator, allowing all agent vehicles to make lane-changing decisions according to the process shown in Figure \ref{Figure 4}.
It is worth noting that to better explore the generality and practicality of the proposed lane-changing decision, we used the Behaviour-Guided Action Prediction (B-GAP) model framework proposed by Angelos et al.\cite{mavrogiannis2022b} instead of the original highway-env model.
Compared to the original highway-env simulator, B-GAP uses the CMetric\cite{chandra2020cmetric} measure to quantify aggressive behaviour such as speeding and overtaking by vehicles. 
It also defines specific driving parameters for aggressive and conservative vehicles within the environment. 
Based on the B-GAP framework, non-intelligent vehicles can exhibit more realistic driving behaviours, breaking away from monotonous rule-based control models, resulting in a more realistic driving environment. 

We constructed a 1000m six-lane motorway environment for training and testing. 
All vehicles were initialised to random lanes with a fixed interval and were travelling at speeds between 20 $km/h$ and 30 $km/h$ at the beginning of the test. The proposed lane-changing decision model \re{is} built in the above environment. 
Due to the similarity in the format of observations and states in this task, both individual Q and global Q \re{adopt} the same network architecture, as shown in \textcolor{black}{Table \ref{table2}}. The parameter settings of the environment and lane change model are shown in \textcolor{black}{Table \ref{table3}}. 

\begin{table}[ht!]   
\begin{center}   
\caption{The architecture of the trajectory prediction network}  
\label{table1} 
\scalebox{0.95}{\begin{tabular}{cc}   
\hline   \textbf{Layer} & \textbf{Parameters} \\   
\hline   GCN Layer 1 & Input size = $V$×2, Output size = 256, activation = $ReLU$  \\ 
   GCN Layer 2 & Input size = 256, Output size = 256, activation = $ReLU$  \\ 
   FC Layer 1 & Input size = 256, Output size = 256, activation = $ReLU$ \\ 
   GRU Layer & Input size = $T_P$×256, Output size = 256, activation = $ReLU$ \\ 
   FC Layer 2 & Input size = 256, Output size = 10, activation = $ReLU$ \\ 
\hline   
\end{tabular}}
\end{center}   
\end{table}

\begin{table}[ht!]   
\begin{center}   
\caption{The architecture of our Lane-change decision network}  
\label{table2} 
\begin{tabular}{cc}   
\hline   \textbf{Layer} & \textbf{Parameters}  \\   
\hline   \makecell{GCN Layer 1 \\ for $h_{sur}$}  & \makecell{Input size = $V$×2(individual Q) or $N$×$V$×2(global Q), \\ Output size = 256, activation = $ReLU$}  \\ 
   \makecell{GCN Layer 2 \\ for $h_{sur}$}  & \makecell{Input size = 256, Output size = 256,\\ activation = $ReLU$}  \\ 
   \makecell{FC Layer 1 \\ for $h_{sur}$}  & \makecell{Input size = 256, Output size = 256,\\ activation = $ReLU$}  \\ 
   \makecell{FC Layer 2 \\ for $h_{env}$}  & \makecell{Input size = 2, Output size = 8,\\ activation = $ReLU$}  \\ 
   \makecell{FC Layer 3 \\ for $h_{ori}$}  & \makecell{Input size = $V$×$F$(individual Q) or $N$×$V$×$F$(global Q), \\ Output size = 256, activation = $ReLU$}  \\  
   \makecell{FC Layer 4 \\ for $h_{ori}$}  & \makecell{Input size = 256, Output size = 256,\\ activation = $ReLU$}  \\ 
   \makecell{FC Layer 5}  & \makecell{Input size = 520, Output size = 5(individual Q) \\ or 125(global Q), activation = $ReLU$}  \\     
\hline   
\end{tabular}   
\end{center}   
\end{table}

\begin{table}[ht!]   
\begin{center}   
\caption{Parameter settings of the environment and the lane-change model}  
\label{table3} 
\begin{tabular}{clc}   
\hline   \textbf{Symbol} & \textbf{Definition} & \textbf{Value} \\   
\hline   $P_D$ & Maximum perception distance & 180m \\ 
   $T_P$ & Maximum of the previous trajectory & 3s  \\ 
   $T_F$ & Minimum of the future trajectory & 1s  \\  
   $v_{min}$ & Highest speed of agent & 1s  \\  
   $v_{max}$ & Lowest speed of agent & 1s  \\  
   $\omega_1$ & Weight of $R_{1}$ & -1  \\  
   $\omega_2$ & Weight of $R_{2}$ & 0.1  \\  
   $\omega_3$ & Weight of $R_{3}$ & 0.4  \\  
   duration & Maximum length for simulation & 40s \\ 
   $lr_{ind}$ & Learning rate for individual value network & 0.0005  \\ 
   $lr_{glo}$ & Learning rate for individual global network & 0.005  \\ 
   $D$ & Experience reply buffer & Size=15000  \\  
   $B$ & Batch of transitions & Size=32  \\  
   $\gamma$ & Discount factor & 0.8  \\  
   $\lambda$ & Weight of the regularization term & 0.3  \\      
   $\alpha$ & Weight of the speed variance term  & 2  \\        
   $\varepsilon$ & Threshold of decision urgency  & 1  \\     
\hline   
\end{tabular}   
\end{center}   
\end{table}

In addition, to assess the adaptability of the proposed method to different environments, we follow the setup of Zhou et al.\cite{zhou2022multi} and create three traffic scenarios with different densities. 
In each scenario, we train and test with different proportions of aggressive vehicles. The specific settings for the traffic scenarios with different densities are shown in \textcolor{black}{table \ref{table4}}.
Our Deep Reinforcement Learning (DRL) training spans a total of 10,000 episodes, with model parameters saved every 1,000 episodes, as well as immediate saving of parameters associated with optimal rewards. The implementation of our method is done using PyTorch. These experiments were conducted on a single Nvidia 2080Ti GPU with an Intel(R) Core(TM) i7-8700k CPU. 

\begin{table}[ht!]   
\begin{center}   
\caption{The settings of the traffic scenarios with different densities}  
\label{table4} 
\begin{tabular}{cccc}   
\hline   \textbf{Traffic density modes} & \textbf{Explanation} & \textbf{AVs} & \textbf{HDVs} \\   
\hline   Sparse & low density traffic patterns & 2 & 8 \\ 
   Normal & middle density traffic patterns & 3 & 15  \\  
   Dense & high density traffic patterns & 5 & 30  \\ 
\hline   
\end{tabular}   
\end{center}   
\end{table}

It is crucial to assess the impact of changing the function of global Q from ``estimating the value of the optimal joint action based on the state" to ``estimating the value of all joint actions based on the state" when configuring the parameters. 
If the global Q does not converge, it not only affects its judgement on joint actions, but also affects the training of the individual Q through the loss shown in Eq.\eqref{11}, which may lead to a failure of decision making even at the individual level. 
During the training of the global Q, its convergence speed may be slower than that of the individual Q due to its decision-making on joint actions with larger action spaces and the limited data collected in each training iteration compared to the parameter-shared individual parts. 
To improve the overall learning efficiency, we adopt the approach of adjusting the learning rates and the dynamic learning rate \cite{yu1995dynamic} to balance the training of individual Q networks for decision making with the acquisition of global knowledge, while accelerating the convergence of the global Q network. 

\begin{figure}[ht!]

	\centering
	\includegraphics[scale=0.45]{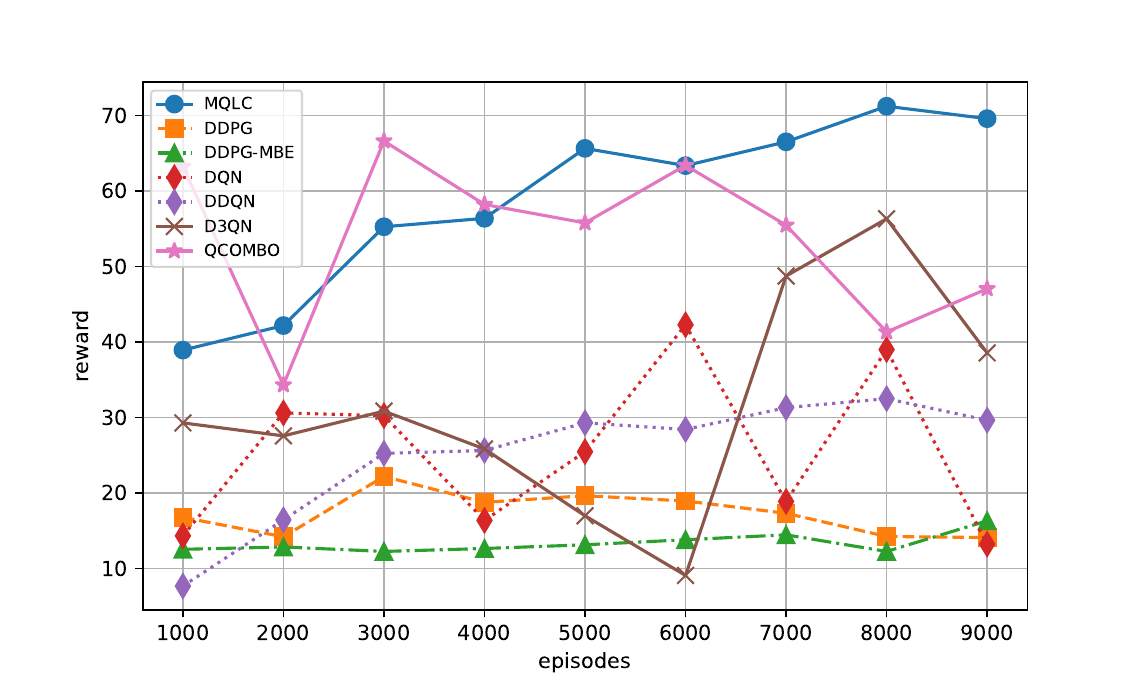}
	\caption{\textcolor{black}{The training process of the comparative experiment. It can be seen from the figure that the ddpg-type method performs the worst. QCOMBO achieves the highest reward effect at the beginning of training, which proves the effectiveness of global decision-making. On this basis, our MQLC method effectively improves the performance due to additional observation acquisition and feature extraction, and reaches the highest level compared with other models in the later stage of training.} }
	\label{Figure 5}
\end{figure}

\begin{table*}[ht!]   
\begin{center}   
\caption{The results of the comparative experiment}  
\label{table5} 
\resizebox{\textwidth}{!}{\begin{tabular}{cccccccccc}
\hline
      \multicolumn{1}{l}{\textbf{Traffic density modes}} & \multicolumn{3}{c}{\textbf{Sparse}} & \multicolumn{3}{c}{\textbf{Normal}} & \multicolumn{3}{c}{\textbf{Dense}} \\
\multicolumn{1}{l}{} &
  \multicolumn{1}{l}{\textbf{Length}} &
  \multicolumn{1}{l}{\textbf{Speed}} &
  \multicolumn{1}{l}{\textbf{Reward}} &
  \multicolumn{1}{l}{\textbf{Length}} &
  \multicolumn{1}{l}{\textbf{Speed}} &
  \multicolumn{1}{l}{\textbf{Reward}} &
  \multicolumn{1}{l}{\textbf{Length}} &
  \multicolumn{1}{l}{\textbf{Speed}} &
  \multicolumn{1}{l}{\textbf{Reward}} \\ \hline
Random    &10.31&24.13&20.99&8.64&23.00&18.33&9.13&21.06&25.71    \\
\textcolor{black}{DDPG\cite{lillicrap2015continuous}} &14.84&23.86&25.60&8.34&23.19&19.29&9.14&22.52&27.35   \\
\textcolor{black}{DDPG-MBE\cite{zhao2023model}} &12.77&23.11&23.42&7.71&22.87&17.62&6.37&22.40&18.29 \\
\textcolor{black}{DQN\cite{mnih2013playing}}       &30.67&20.69&42.50&22.22&23.03&49.47&18.32&19.70&48.52\\
\textcolor{black}{Double DQN\cite{van2016deep, 8951626}}&33.28&24.13&\textbf{43.21}&29.93&21.86&63.09&23.60&20.94&67.49\\
\textcolor{black}{D3QN\cite{han2020research, peng2021end} }     &28.76&23.69&39.78&31.84&22.49&67.72&23.08&22.97&65.74\\
\textcolor{black}{QCOMBO\cite{zhang2019integrating}} &25.04&\textbf{24.25}&34.11&29.18&\textbf{25.60}&66.23&25.76&22.37&75.52\\
MQLC      &\textbf{33.86}&23.80&42.89&\textbf{34.89}&24.94&\textbf{71.52}&\textbf{28.81}&\textbf{25.08}&\textbf{85.77}\\ \hline
\end{tabular}}
\end{center}   
\end{table*}

\subsection{Baseline methods and Evaluation Metrics}
\textcolor{black}{Our experiments are conducted in B-GAP\cite{mavrogiannis2022b} environment. We select the following baseline models for comparative experiments in order to fully validate the effectiveness of our proposed method:}

\textcolor{black}{
\begin{itemize}[leftmargin=2em] % 调整左边距缩进
    \item \textbf{Random} adopt random actions as the agent's decision.
    \item \textbf{DDPG}\cite{lillicrap2015continuous} combines the benefits of DQN (Deep Q-Network) and Policy Gradient methods, handling continuous action spaces effectively.
    \item \textbf{DDPG-MBE}\cite{zhao2023model} enhances DDPG by generating additional training data through simulated rollouts.
    \item \textbf{DQN}\cite{mnih2013playing} is an algorithm that combines Q-learning with neural networks. It uses a neural network to approximate the Q-value function, which estimates the expected rewards for state-action pairs.
    \item \textbf{Double DQN}\cite{van2016deep, 8951626} uses two networks to decouple action selection from action evaluation, leading to more accurate Q-value estimates which results in improved stability and performance in training. 
    \item \textbf{D3QN}\cite{han2020research, peng2021end} combines Dueling Network Architectures with Double DQN. It separates the estimation of the state value and the advantage for each action, which helps in better learning of state values and more accurate Q-value estimates. In the experiment, we set its input as state to ensure the fairness of the comparison experiment.
    \item \textbf{QCOMBO}\cite{zhang2019integrating} is a multi-agent reinforcement learning algorithm for optimizing traffic signal networks. It combines independent and centralized learning using a consistency regularizer to align individual and global action-value functions. We adapt it to the lane-changing decision environment.
\end{itemize}
}

\textcolor{black}{Among the above models, the architecture and design of DQN and its variants Double DQN and D3QN are all from the original B-GAP\cite{mavrogiannis2022b} environment. Their parameters also adopt the original parameters set by the environment.}

\textcolor{black}{As for the evaluation metrics, we emphasise the consideration of the reward component in the evaluation metrics, as rewards serve as the only feedback signal during reinforcement learning training. 
They directly reflect the quality of the actions taken by the intelligent agent. 
% Based on the specific definition of the reward function proposed in \re{chapter \ref{sec:3.1}}, 
\re{The} efficiency and safety of lane-changing decisions are also the key objectives we consider. The average speed of agents provides a clear understanding of the benefits of their decisions and can also indirectly reflect the overall traffic efficiency.
We therefore choose it as a measure of efficiency. 
And in the simulation environment, collisions lead to the termination of episodes before the maximum episode length is reached. The longer the episode length is, the safer the lane-changing decision made by the agent is.
As a result, the average episode length serves as an indicator for evaluating the safety of the models.}

\subsection{Result of comparative experiments}
\label{sec:4.2}
To demonstrate the superiority of the proposed MQLC in decision making, we conduct comparative experiments.
Comparative experiments were carried out in the three traffic environments shown in \textcolor{black}{Table \ref{table4}}. The ``Sparse" mode in the table represents a low-density traffic pattern where there are two agents and 8 surrounding vehicles; the ``Normal" mode represents a medium-density traffic pattern with 3 agents and 15 surrounding vehicles; and the ``Dense" mode represents a high-density traffic pattern with 5 agents and 30 surrounding vehicles. 

\textcolor{black}{Figure \ref{Figure 5} shows the training process of the comparative experiment. From the figure, we can clearly see that the DDPG and DDPG-MBE methods show the worst performance, which proves that this type of method is not suitable for discrete action spaces. Compared with the benchmark DQN, Double DQN and D3QN models, QCOMBO shows outstanding performance. This is because the global decision-making guidance feature of QCOMBO allows the agent to have more considerations for the environment, thereby improving the overall lane-changing effect. On this basis, the MQLC model proposed in this paper optimizes observation acquisition and feature extraction, innovatively adopts a global decision-making method, and achieves the best effect in the later stage of training. This figure further supplements the effectiveness of the method proposed in this paper.}

\textcolor{black}{Table \ref{table5}} shows the results of the comparative experiments in terms of safety, efficiency and reward metrics. 
From \textcolor{black}{Table \ref{table5}}, we can draw the following conclusions:

(1) \textcolor{black}{Comparing the ``MQLC" row in Table \ref{table5} with the ``DDPG" and ``DDPG-MBE" rows, it can be seen that both models lag far behind our method in terms of average episode duration and total reward indicators. 
\re{For example, MQLC outperforms DDPG-MBE by $305.9\%$ in
terms of ``Reward" in the ``Normal" traffic scenario. The difference in this metric between the two methods is obvious.}
The reason why the DDPG method lags behind is that it is proposed to solve the problem of continuous action space, while the lane change decision task belongs to the discrete action space problem with fewer discrete points. The large difference in the action space between the two leads to inadaptability. This \re{indicates} that even the most advanced multi-agent collaboration method may not be applicable to lane-changing decision tasks due to the particularity of the decision-making environment and settings.}

(2) Comparing the ``MQLC" row in \textcolor{black}{Table \ref{table5}} with the ``DQN", ``Double DQN" and ``D3QN" rows, it can be observed that MQLC achieves the best performance among the  reinforcement learning methods. In the ``Normal" and ``Dense" traffic scenarios, MQLC achieves the highest average episode length, the highest speed, and the optimal reward. 
\re{For instance, MQLC outperforms D3QN by $5.6\%$ and $30.5\%$ in
terms of ``Reward" in the ``Normal" and ``Dense" traffic scenarios.}
However, in the ``Sparse" scenario, all models show similar performance and MQLC's performance is not \re{evidently} better than the other models. This is because MQLC's primary innovation lies in the coordination between multiple intelligent agents. When there are relatively few agents, MQLC does not show a clear advantage. However, as the number of agents increases, MQLC can optimise the overall traffic flow by using additional feature extraction and coordinated decision making, balancing individual and global interests, and thus achieving optimisation of the overall traffic flow. 

(3) Comparing the ``MQLC" row with the ``QCOMBO" row shows that MQLC outperforms QCOMBO in various scenarios \re{in general}. 
\re{In terms of ``Length" which evaluates the safety, MQLC outperforms QCOMBO by $35.2\%$, $19.6\%$ and $11.8\%$  in the ``Sparse", ``Normal" and ``Dense" traffic scenarios.}
This is because QCOMBO only considers bringing individual network estimates closer to the global network during training, but continues to use individual decisions during actual decision making, and is unable to overcome the limitations of information acquisition and decision coordination. In contrast, MQLC fully exploits the advantages of CVIS, uses state information to understand the overall traffic situation, and effectively coordinates during decision making by using individual agents to support global decision making. As a result, MQLC achieves significantly better lane-changing decision performance than QCOMBO.

(4) Comparing the three performance metrics across various scenarios, MQLC consistently outperforms other models in terms of ``Length" and ``Reward" in nearly all cases. However, except for the most congested ``Dense" scenario, there isn't a significant advantage in the ``Speed" metric in the other two scenarios. These results suggest that MQLC places a higher emphasis on prioritizing the safety of all agents' actions through collaborative efforts, thereby avoiding collision penalties and ultimately achieving superior lane-changing decisions, rather than solely focusing on efficiency enhancement.

\begin{table}[ht!]   
\begin{center}   
\caption{The effect of architecture of intent prediction network}  
\label{table9} 
\begin{tabular}{cccc}
\hline
 Architecture            & Episodes length & Average speed & Total reward \\ \hline
LSTM   &    32.87          &       24.30        &       69.49       \\
GCN+GRU &       35.08          &       24.34        &       71.22     \\ \hline
\end{tabular}
\end{center}   
\end{table}
\vspace{-20pt}

\textcolor{black}{\subsection{Design of the intent prediction network}
In \re{section \ref{sec:3.1}}, we introduced the design of the intention prediction part in detail, that is, to expand the observation by predicting the trajectories of vehicles around the agent to improve the ability of the agent to collaborate. In this section, we analyze the specific design of the trajectory prediction network in table \ref{table1}. For comparison, we implemented the LSTM-based intention prediction network under “normal” traffic density conditions while keeping other conditions the same. The experimental results are shown in table \ref{table9}.}

\textcolor{black}{From the analysis, we can see that the final lane-changing effect of the intention prediction network based on GCN combined with GRU adopted in this paper is much better than that of the LSTM-based model in terms of episodes’ average length and comprehensive reward. This is because in the lane-changing scenario, the interaction between vehicles can be perfectly represented by the topological structure. The GCN-based model can accurately capture the interaction relationship between vehicles in the lane-changing decision, and then use the more computationally efficient GRU model for time series prediction, more accurately predicting the future trajectory of surrounding vehicles and providing a reference for decision-making. This result also proves the superiority of the architecture selected in this paper.}

\begin{table}[ht!]   
\begin{center}   
\caption{The effect of $\lambda$ in performance}  
\label{table6} 
\begin{tabular}{cccc}
\hline
             & Episodes length & Average speed & Total reward \\ \hline
$\lambda$=0.1 &        30.49         &       24.21        &        66.98       \\
$\lambda$=0.3 &        35.08         &       24.34        &        71.22      \\
$\lambda$=0.5 &        30.60         &       23.85        &        65.19      \\ \hline
\end{tabular}
\end{center}   
\end{table}
\vspace{-10pt}

\begin{figure}[ht!]
\centering
	\subfloat[The effect of $\lambda$ in training process.]{
	\centering
	\includegraphics[scale=0.25]{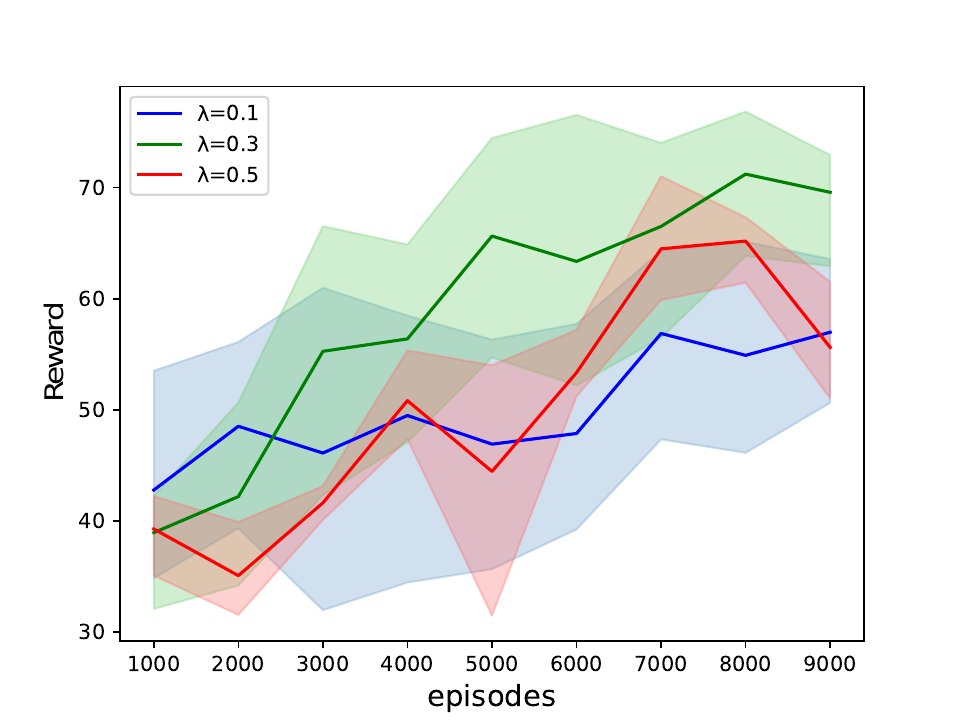}}    
\subfloat[The effect of $\varepsilon$ in training process.]{
\centering
\includegraphics[scale=0.25]{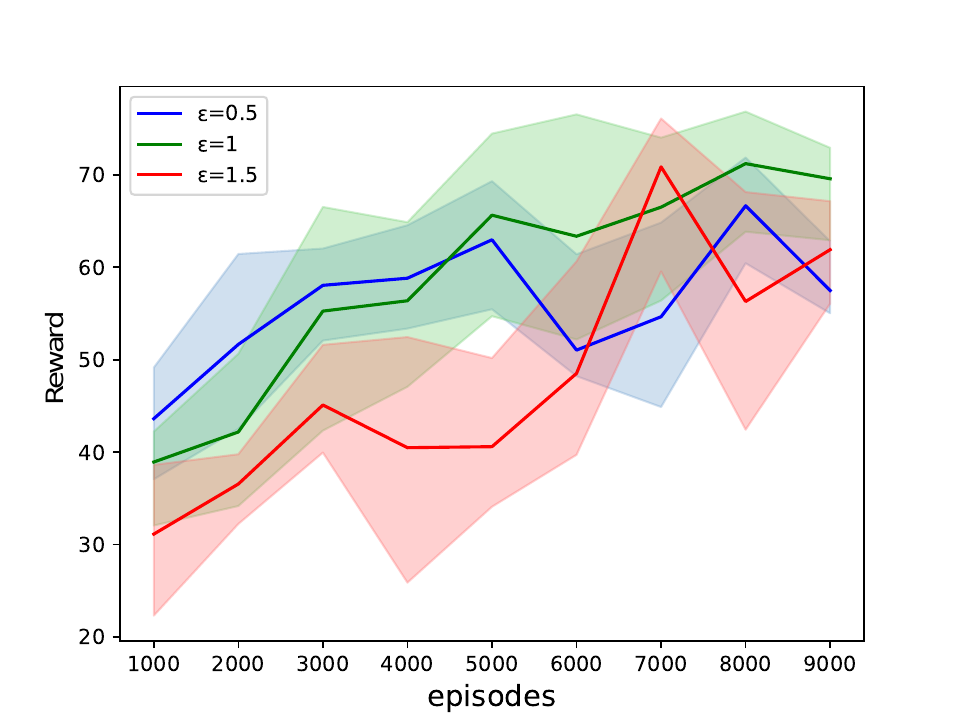}
}
	\caption{The three coloured lines or shaded regions represent the performance of the MQLC model in lane-changing decisions over five training runs. In the graph, the lines represent the average reward obtained in lane-changing decisions over the corresponding training episodes in the five runs. The corresponding shaded regions represent the range of rewards achieved by models with the corresponding $\lambda$ or $\varepsilon$ values during the corresponding training episodes. Both in terms of lines and regions, the model with $\lambda$ = 0.3 and $\varepsilon$ = 1 consistently shows the best performance.}
	\label{Figure 6}
\end{figure}

\begin{table}[ht!]   
\begin{center}   
\caption{The effect of $\varepsilon$ in performance}  
\label{table7} 
\begin{tabular}{cccc}
\hline
             & Episodes length & Average speed & Total reward \\ \hline
$\varepsilon$=0.5 &       29.97          &       26.11        &       64.28       \\
$\varepsilon$=1 &       35.08          &       24.34        &       71.22     \\
$\varepsilon$=1.5 &       33.85          &       25.34        &       71.30       \\ \hline
\end{tabular}
\end{center}   
\end{table}

\begin{figure}[ht!]
\centering
\subfloat[Histogram of urgency.]{
	\centering
	\includegraphics[scale=0.25]{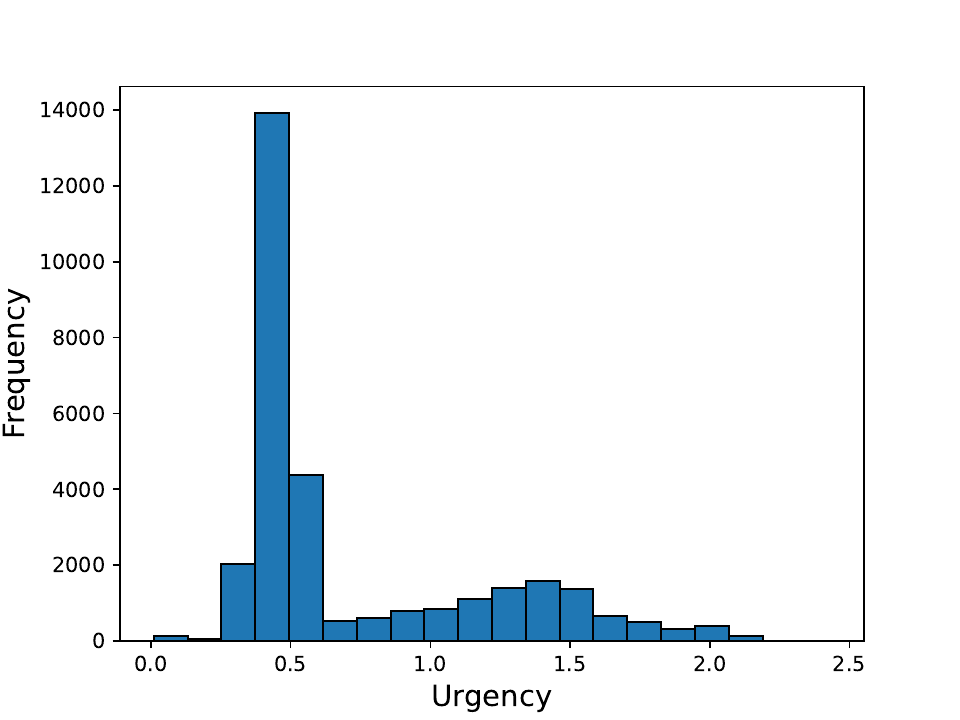}}    
\subfloat[Pie chart of urgency.]{
\centering
\includegraphics[scale=0.25]{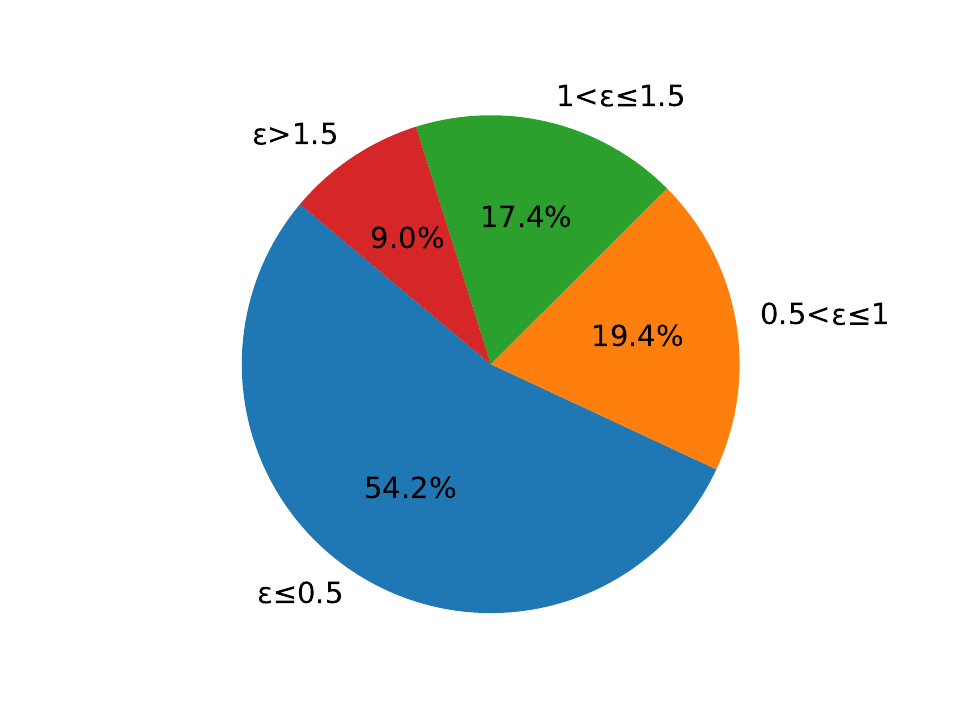}
}
	\caption{The frequency of urgency during a training process. Figure \ref{Figure 7}(a) shows the distribution of urgency values of intelligent agents on all decision instances collected during a single training run. On the other hand, Figure \ref{Figure 7}(b) shows the proportion of the frequency of these urgency values within different intervals.}
	\label{Figure 7}
\end{figure}

\subsection{Parameter settings of training status}
\label{sec:4.3}
In multi-agent tasks, achieving high-performing models during training is challenging due to the instability of policy updates and the complexity of exploration. 

To address the former, we use a probability-based sampling approach that allows agents to choose from a wider range of actions to avoid being trapped in local optima. Specifically, after the value network computes the expected payoffs for each action, we update the action sampling strategy, increasing the probability that agents choose actions with higher expected payoffs. This randomness is gradually reduced as training progresses.

Regarding the latter concern, lane-changing vehicles are homogeneous agents sharing a value network, which causes instability mainly due to the consistency regularisation parameter $\lambda$. 
In addition to the aforementioned challenges, the reliability of priority settings, which serve as the basis for joint decision making in MQLC, requires further investigation. We explore the rationality of these two types of parameter configurations - the consistency regularising parameter $\lambda$ and the urgency threshold $\varepsilon$ - both from experimental results and from the training process. 
All experiments in this subsection are conducted under the 'normal' traffic density condition.
\subsubsection{The consistency regularisation parameter $\lambda$}
\label{sec:4.3.1}
As mentioned in chapter \ref{chap:3.2.1}, the setting of the lambda parameter affects the degree to which the individual Q methods influence the global Q, thus affecting both the experimental results and the convergence efficiency of the network. Therefore, its configuration is crucial in the experiments. 
\textcolor{black}{Table \ref{table6}} and Figure \ref{Figure 6}(a) show the performance and training process of MQLC models with different $\lambda$ configurations, respectively, while maintaining uniform conditions, under the 'normal' decision scenario. 

From the experimental results, it can be seen that the MQLC performs optimally when $\lambda$ is set to 0.3, both in terms of training performance and decision results. Looking at the training curves, models with lower $\lambda$ values show better initial performance, but their final performance falls short of the model with a $\lambda$ value of 0.3. We attribute this phenomenon to the fact that models with lower $\lambda$ values tend to overlook global considerations to some extent. While they may make effective decisions based on faster convergence of individual Q-networks during early stages of training, the lack of collaborative thinking ultimately limits their decision performance.
On the other hand, when $\lambda$ values are set to higher values, the model shows consistently poorer performance. There are two main reasons for this. First, models with higher $\lambda$ values overemphasise the global impact of lane-changing decisions. Conversely, agents that rely solely on their own observations as decision criteria struggle to make appropriate choices that take into account both global interests and their individual situations. Second, based on the analysis of the differences in training efficiency between the two networks, an excessively high $\lambda$ value interferes with the training of individual Q-networks by the global Q-network, rendering agents unable to accurately analyse the environment.

\begin{figure}[ht!]
	\centering
	\includegraphics[scale=0.35]{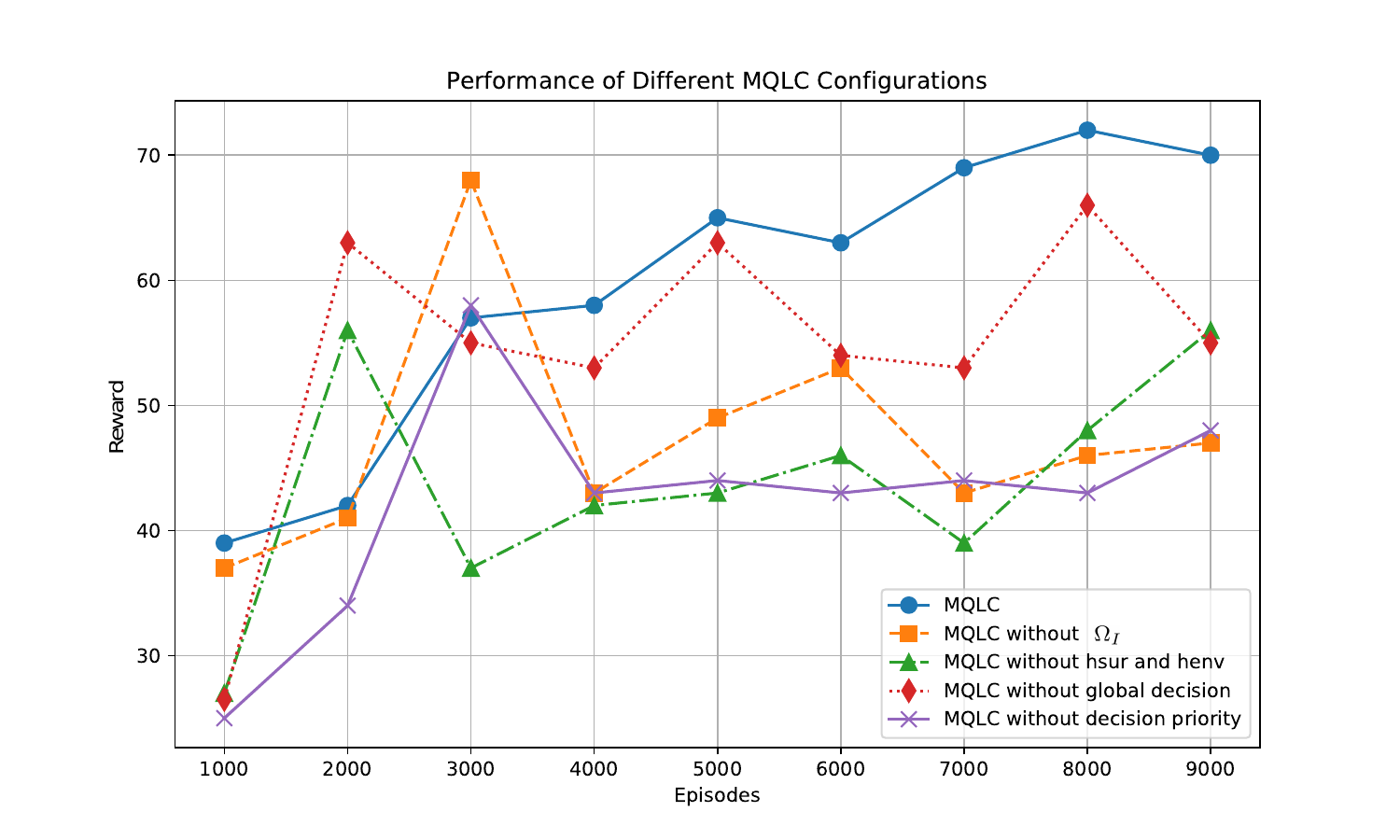}
	\caption{The training progress of the ablation experiment. The graph shows the average performance of the original MQLC model and models with specific components ablated over five training runs for lane-change decisions. In the graph, 'MQLC' represents the performance of the original model. MQLC without $\Omega_I$' represents the model with the intention prediction component removed from the observations. MQLC without $h_{sur}$ and $h_{env}$' represents the model with the additional feature extraction structure removed from the Q-net architecture. MQLC without global decision' indicates the model with the global decision process removed, relying only on individual agent models for decision making. MQLC without decision priority' denotes the model with the concept of decision priorities removed, resulting in all agents strictly adhering to the global decision model.}
	\label{Figure 8}
\end{figure}

\begin{table}[ht!]   
\begin{center}   
\caption{The results of the ablation experiment}  
\label{table8} 
\scalebox{0.9}{\begin{tabular}{ccccc}
\hline
\multicolumn{2}{c}{}                        & Episodes length & Average speed & Total reward \\ \hline
\multicolumn{2}{c}{\textbf{MQLC}}           & 35.08           & 24.34         & 71.22        \\ \cline{1-2}
\multicolumn{1}{c|}{\multirow{4}{*}{\begin{tabular}[c]{@{}c@{}}MQLC \\ without\end{tabular}}} & $\Omega_I$ & 29.7 & 23.48 & 61.77 \\
\multicolumn{1}{c|}{} & $h_{sur}$ and $h_{env}$ & 26.5            & 25.55         & 59.22        \\
\multicolumn{1}{c|}{} & global decision     & 26.84           & 24.92         & 58.49        \\
\multicolumn{1}{c|}{} & decision priority   & 20.99           & 25.17         & 39.96        \\ \hline
\end{tabular}}
\end{center}   
\end{table}

\subsubsection{The urgency threshold $\varepsilon$}
\label{sec:4.3.2}
By using equation \ref{13} and equation \ref{14}, we achieve the classification of agents' decision priorities by setting the urgency threshold $\varepsilon$. 
This classification serves as the basis for facilitating lane-changing decisions that benefit both parties. 
Therefore, the configuration of $\varepsilon$ has a significant impact on the decision making process. 

To gain a clearer understanding of the specific design of $\varepsilon$, we collected the urgency indicators for all intelligent agents during decision making over a 1000-step training process. We then visualise this data as shown in Figure \ref{Figure 7}.

As can be seen from the graph above, more than half of the decision instances show that the urgency $\varepsilon$ for the intelligent agents is less than 0.5. In addition, 9\% of the instances have $\varepsilon$ values greater than 1.5, while the remaining instances are almost evenly distributed between the remaining two intervals. We ran tests on the lane-changing performance of models with $\varepsilon$ values of 0.5, 1 and 1.5. The results are shown in \textcolor{black}{table \ref{table7}} and figure \ref{Figure 6}(b).

From the graphs and plots of the experimental results, it can be seen that $\varepsilon$=1 shows relatively better training and testing performance. Although $\varepsilon$=1.5 and $\varepsilon$=1 have similar optimal reward results, the performance of $\varepsilon$=1.5 is characterised by higher variance and slower convergence in the training process, leading to less ideal training results. This can be attributed to the fact that at higher values of $\varepsilon$ very few decision instances are considered urgent, and as a result the global Q-network is heavily involved in decision making. Due to its limitations in convergence and individual specificity, the global Q-network tends to make decisions that benefit the overall gain but may compromise the interests of individual agents. While this approach may lead to coordination in some cases, it often disrupts normal traffic operations for most lane-changing decisions.
When $\varepsilon$ is set to a lower value, more decision instances are deemed urgent without requiring coordination from the global Q-network. However, this approach to decision making is not conducive to inter-agent cooperation and ultimately reduces overall traffic efficiency.

% \begin{figure}[ht!]
% 	\centering
% 	\includegraphics[scale=0.7]{fig5.jpeg}
% 	\caption{ The three colored lines and shaded regions respectively represent the performance of the MQLC model with $\varepsilon$ values of 0.5, 1, and 1.5 during lane-changing decisions across five training runs. In the graph, the lines depict the average reward obtained in lane-changing decisions over the corresponding training episodes in the five runs. The corresponding shaded regions represent the range of rewards achieved by models with the respective $\varepsilon$ values during the corresponding training episodes. Both from the perspective of the lines and the regions, the model with $\varepsilon$ = 1 consistently demonstrates the best performance.}
% 	\label{Figure 6}
% \end{figure}

\begin{figure*}[ht!]
\centering
\subfloat[Traffic situation(1) at time $t-2$.]{
\includegraphics[scale=0.25]{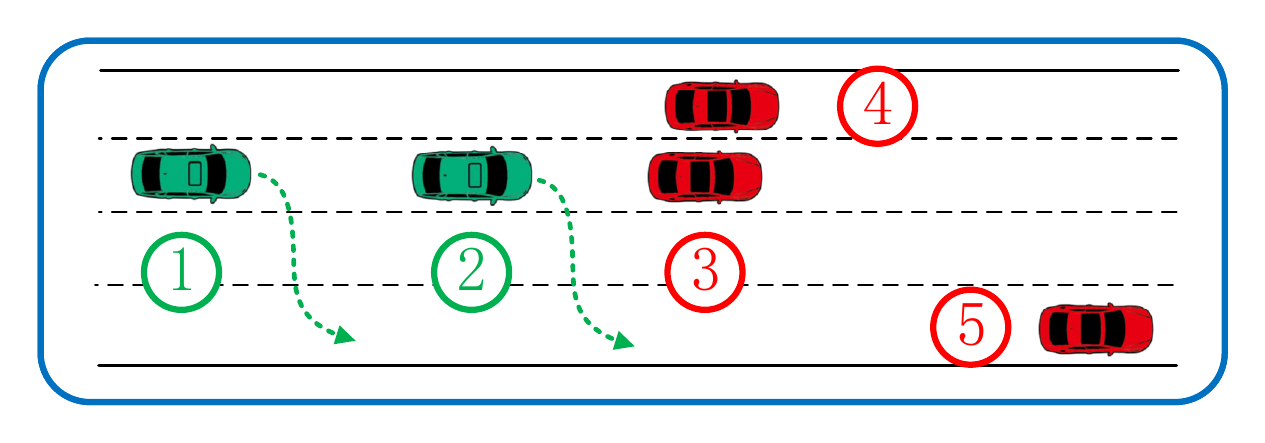}
}
\subfloat[Traffic situation(2) at time $t-2$.]{
\includegraphics[scale=0.25]{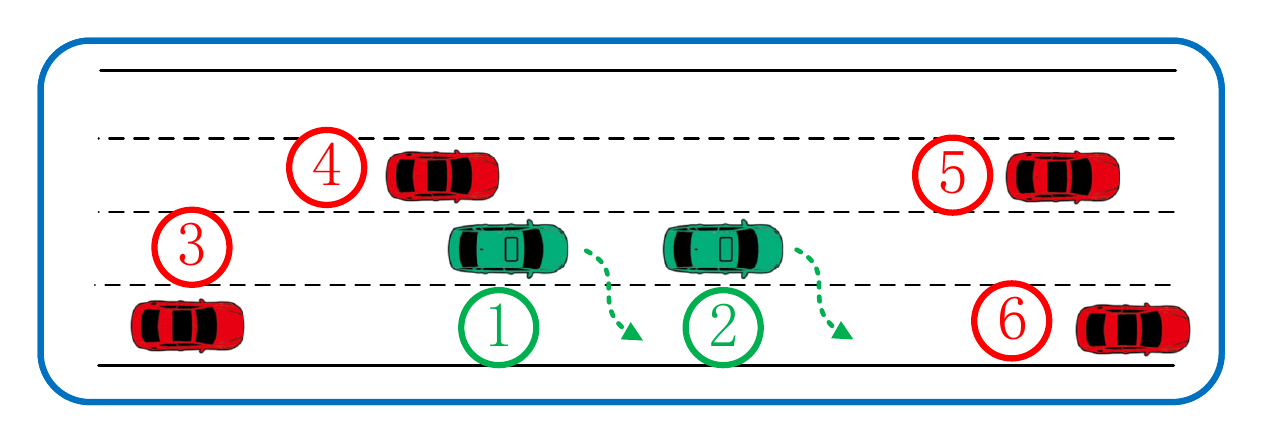}
}
\subfloat[Traffic situation(3) at time $t-2$.]{
\includegraphics[scale=0.25]{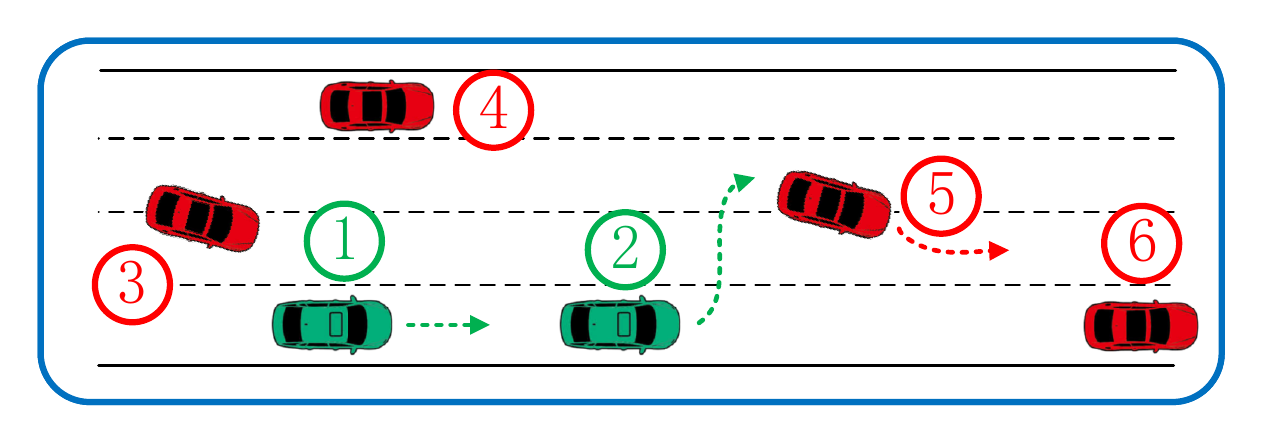}
}
\quad
\subfloat[Traffic situation(1) at time $t-1$.]{
\includegraphics[scale=0.25]{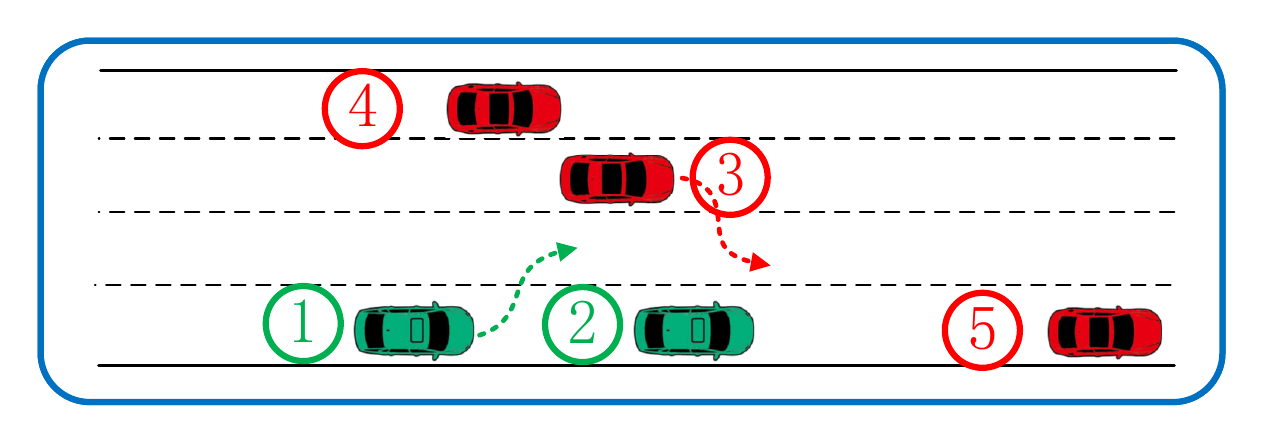}
}
\subfloat[Traffic situation(2) at time $t-1$.]{
\includegraphics[scale=0.25]{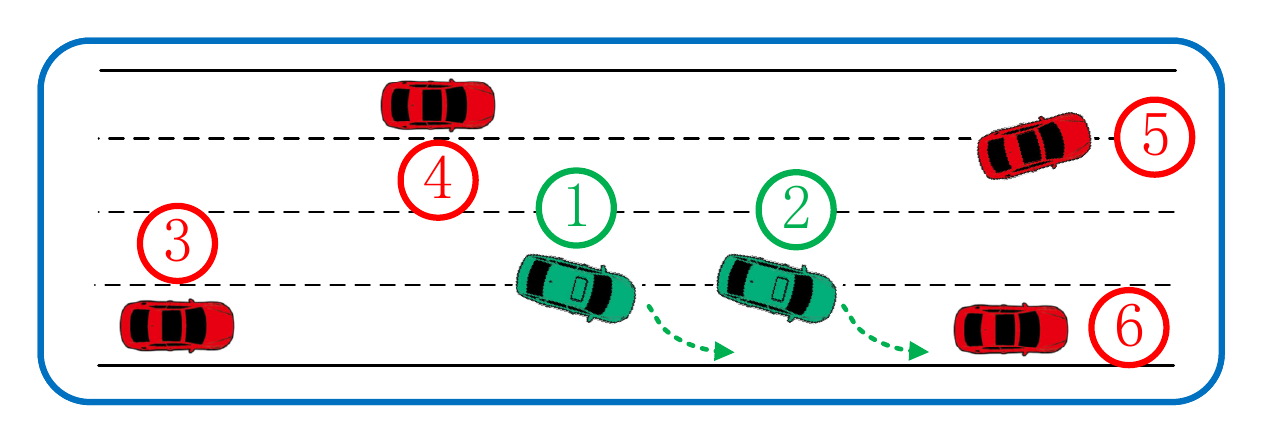}
}
\subfloat[Traffic situation(3) at time $t-1$.]{
\includegraphics[scale=0.25]{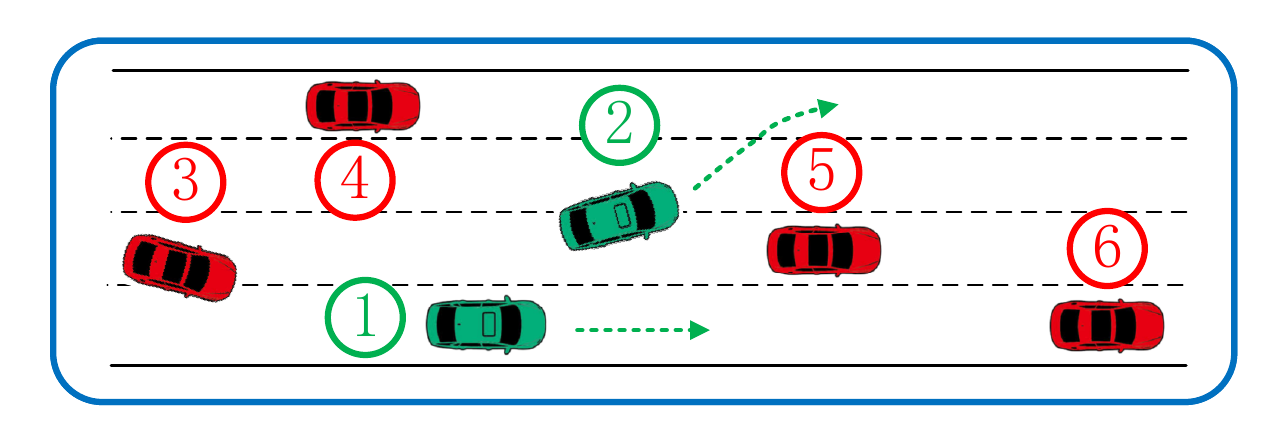}
}
\quad
\subfloat[Traffic situation(1) at time $t$.]{
\includegraphics[scale=0.25]{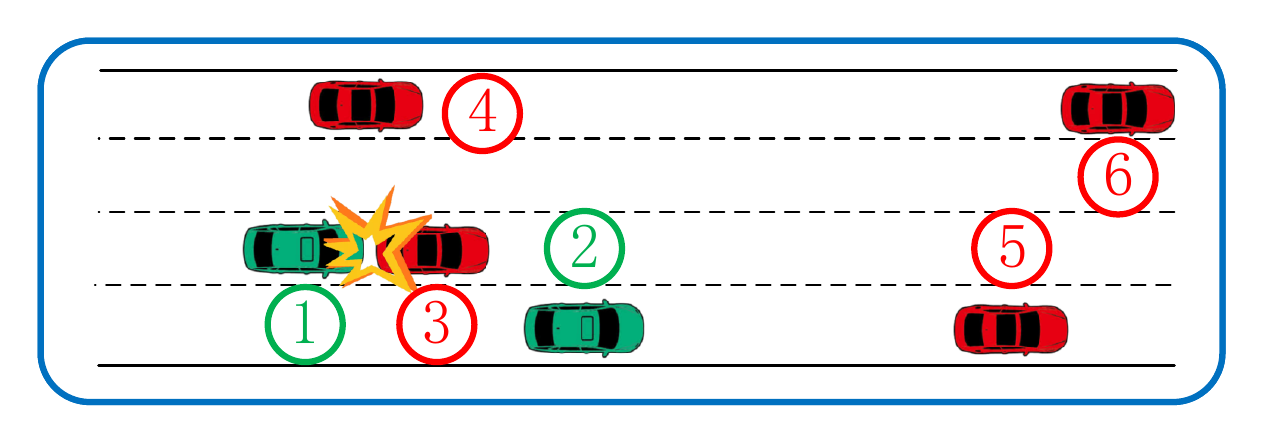}
}
\subfloat[Traffic situation(2) at time $t$.]{
\includegraphics[scale=0.25]{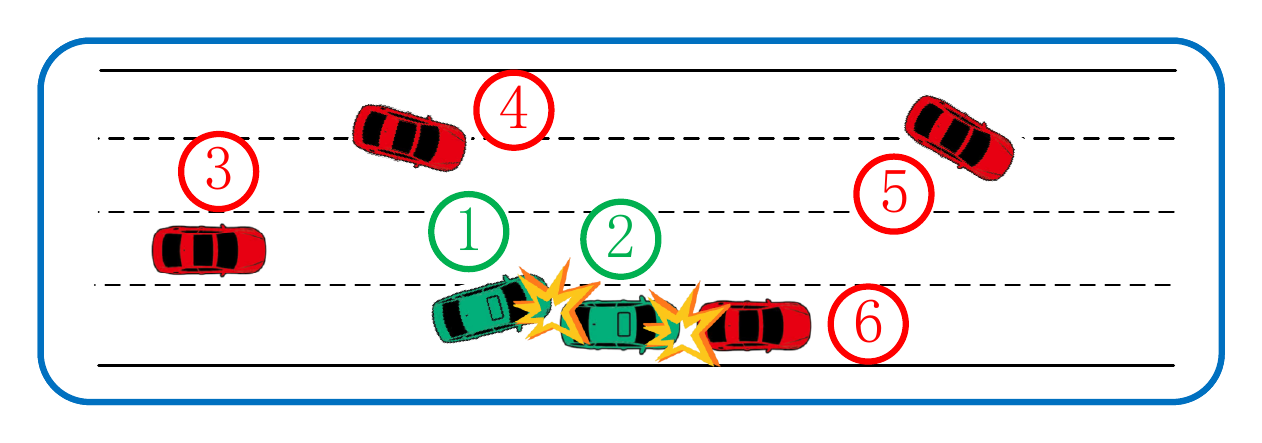}
}
\subfloat[Traffic situation(3) at time $t$.]{
\includegraphics[scale=0.25]{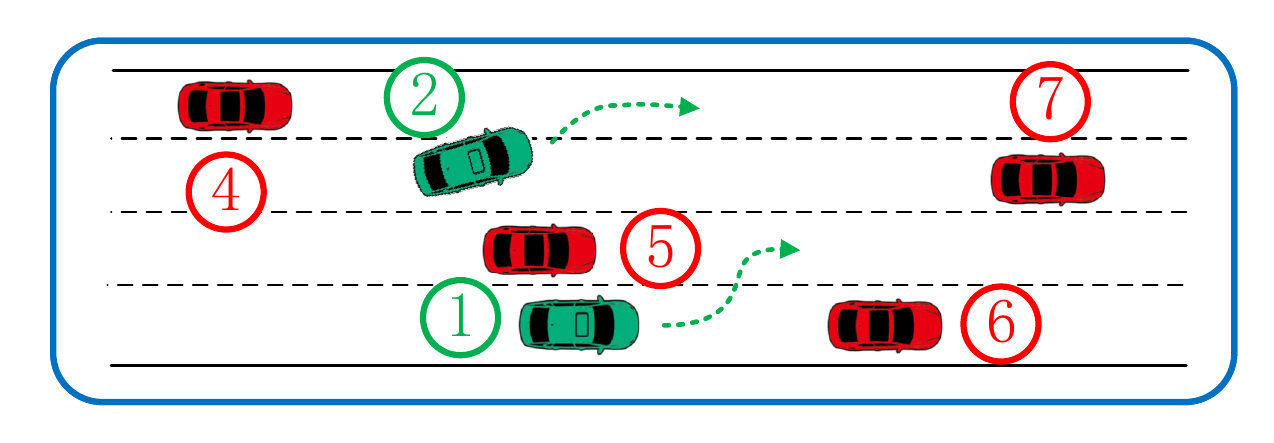}
}
\caption{The results of the visualisation experiment. The situation (1) and (2) are for MQLC without global decision and the situation (3) is for MQLC. The situation (1) and (2) illustrate two accident scenarios resulting from multi-agent lane-changing decisions without global coordination respectively. In situation (1) at time $t-2$, due to the obstruction caused by vehicles 3 and 4 to intelligent agents 1 and 2, both agents decide to change lanes to the right in order to obtain better driving conditions. At time $t-1$, both agents reach their desired positions, but agent 2 inadvertently obstructs agent 1, forcing it to change lanes to the left, ultimately leading to a collision between vehicles 1 and 3 at time $t$. In situation (2) at time $t-2$, both agents decide to change lanes to the right for better driving conditions. At time $t$, agent 2 rear-ends vehicle 6 as a lack of coordination prevents it from maintaining a safe following distance. Subsequently, agent 1 also rear-ends agent 2, resulting in a serious accident. The situation (3) shows the excellent performance of global decision making in a complex traffic scenario. At time $t-2$, as vehicle 5 changes lanes to the right, driving conditions deteriorate for both agents. Seeking better driving conditions, Agent 2 proactively changes lanes to the left twice, creating space for Agent 1 and improving its own driving visibility. At time $t$, both agents have achieved improved driving conditions.}
\label{Figure 9}
\end{figure*}

% \begin{figure}[ht!]
% \centering
% \subfloat[Traffic situation at time $t-2$.]{
% \includegraphics[scale=0.4]{fig9g.pdf} \label{8g}
% }
% \quad
% \subfloat[Traffic situation at time $t-1$.]{
% \includegraphics[scale=0.4]{fig9h.pdf} \label{8h} 
% }
% \quad
% \subfloat[Traffic situation at time $t$.]{
% \includegraphics[scale=0.4]{fig9i.pdf}\label{8i}
% }
% \caption{The results of the visualisation experiment for MQLC. This figure shows the excellent performance of global decision making in a complex traffic scenario. At time $t-2$, as vehicle 5 changes lanes to the right, driving conditions deteriorate for both agents. Seeking better driving conditions, Agent 2 proactively changes lanes to the left twice, creating space for Agent 1 and improving its own driving visibility. At time $t$, both agents have achieved improved driving conditions.}
% \label{Figure 10}
% \end{figure}

\subsection{Ablation experiment}
\label{sec:4.4}
In order to visually illustrate the specific impact of each innovation on decision making, in this subsection we have conducted ablation experiments on different innovations within MQLC. 
All experiments in this subsection are conducted under the 'normal' traffic density condition. 
The first two components were achieved by removing the relevant parts of the observations and networks from the original model. The third component involved transforming the decision making approach for all agents to be solely determined by individual Q-networks, in order to investigate the impact of the absence of the global decision making component on the experimental results. The final component consisted of configuring all intelligent agents to be controlled solely by the global Q-network, without considering individual agent specifics.
These experiments included intention prediction, decision network architecture, global Q involvement, and the concept of decision priorities. 
\textcolor{black}{The rewards and final results of each model training stage in the experiment are shown in the figure \ref{Figure 8} and Table \ref{table8} respectively.}

\textcolor{black}{Analysis of Figure \ref{Figure 8} shows that compared with the models in the ablation experiment, MQLC shows the best training effect. The two models that ablated observation and network improvement showed good results in the early stage of training, but due to insufficient input information or feature extraction capabilities, the effect was not as good as MQLC in the later stage. The model that ablated global decision or priority showed unreasonable lane-changing effects in the entire stage of training, which shows that the above two aspects are the basis for MQLC to achieve global coordination. No matter which one is missing, the coordination cannot proceed normally.}

The comparison between models without $\Omega_I$ and the original model shows that the decision performance of the MQLC model deteriorates significantly when the intention prediction component is missing from the agents' observations. This degradation is attributed to the fact that intention prediction provides additional useful information to the agents, allowing them to learn the driving patterns of the surrounding vehicles. This allows the agents to implicitly incorporate cooperation with other vehicles into their decisions. This underlines the importance of intention prediction as a fundamental element in achieving collaborative decision making.
The comparison between models without $h_{sur}$ and $h_{env}$ and the original model highlights that our improved network, enriched with additional network structures, provides enhanced decision information. This modification proves to be more conducive for the intelligent agents to make rational lane-changing decisions.
The MQLC without global decision making, which keeps all other aspects unchanged while adopting the decision making approach of CTDE, initially showed rapid convergence and made the most efficient decisions. However, as training progressed, it gradually fell behind the performance of the original model. This shows that the global decision approach sets a higher bound for the agents, which serves as a basis for decision cooperation.
The MQLC model without decision priorities showed the worst performance of all the models compared. The reason for this is that complete reliance on global decision making effectively eliminates consideration of the specific circumstances of individual agents. Consequently, certain agents in disadvantaged positions are forced to adhere to global interests, potentially leading to the formulation of risky decisions. The performance of this model suggests that over-reliance on global coordination-based decision making is not conducive to cooperation among intelligent agents in lane-changing decisions.

\subsection{The visualization experiment}
\label{sec:4.5}
Here, we conduct a visualisation experiment to qualitatively analyse the impact of the global decision strategy on lane-changing outcomes by comparing the actual performance of MQLC with and without the adoption of the global decision strategy. \textcolor{black}{To facilitate observation, the experiment is conducted in a simulated environment on a sufficiently long four-lane highway.} This environment consists of three special agents and five surrounding vehicles with aggressive driving styles. 
Figures \ref{Figure 9} shows the decision situation of the agent. \textcolor{black}{The situation depicted in the figure is a reconstruction of an actual scenario that occurred within the highway-env simulation environment.}
From the comparison, it can be seen that global decision making provides a collaborative prerequisite for the intelligent agent to make lane-changing decisions. Through global decision making, the lane-changing decision made by the agent can consider other agents while ensuring its own interests, and continuously improve the safety and efficiency of driving through mutual cooperation. Only through global decision making can intelligent agents overcome the limitations of individual observation and achieve excellent lane-changing decisions. 

\section{Conclusion and future work}
\textcolor{black}{In this paper, to the best knowledge, we propose the first collaborative multi-agent lane-changing decision framework named \textbf{M}ix \textbf{Q}-learning for \textbf{L}ane \textbf{C}hanging (MQLC) to improve the performance of lane changing decisions through collaboration of agents.} At the individual level, MQLC enhances decision information through intent recognition based on trajectory prediction, adapts the network structure to the characteristics of decision information, and enables more comprehensive feature extraction. At the cooperative level, varying degrees of decision autonomy are granted to the agents according to the urgency of their decisions. Agents make decisions based on individual optimal value functions. Subsequently, the global Q-network coordinates feasible individual decisions using global information to achieve a balance between individual interests and overall benefits. Extensive experimental results on various complex traffic scenarios show that our proposed MQLC model outperforms other state-of-the-art multi-agent lane-changing decision methods in terms of efficiency, safety, and other metrics. The experiments also highlight MQLC's approach of global coordination of individual decisions has potential applications in other multi-agent reinforcement learning tasks, providing valuable insights into multi-agent collaboration. 
However, it is worth noting that our collaborative approach focuses primarily on increasing overall reward and does not consider other attributes of vehicle motion, such as speed and comfort. In future work, there is scope to explore more diverse reward criteria to improve the model's performance on intuitive motion attributes.

\bibliographystyle{IEEEtran}
\bibliography{IEEEabrv,reference}
\begin{IEEEbiography}
[{\includegraphics[width=1in,height=1.25in,clip,keepaspectratio]{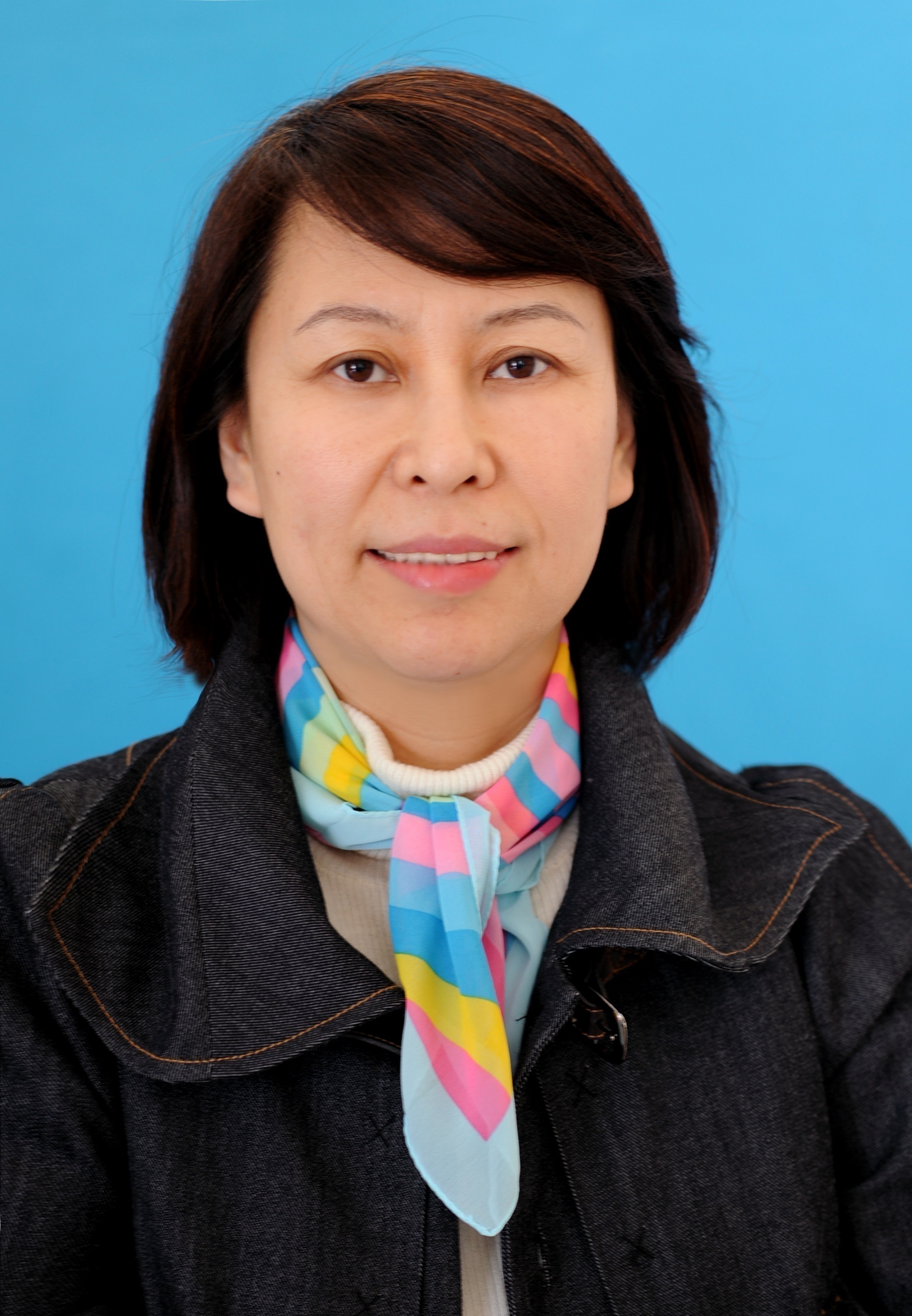}}]{Xiaojun Bi}
% received the M.S. degree from Harbin Institute of Technology, Harbin, China, in 1990, and the Ph.D. degree from Harbin Engineering University, Harbin, China, in 2006. She is a professor and Ph.D. supervisor at college of Information And Communication Engineering, Harbin Engineering University, Harbin, China. Her research interests include evolutionary computation, multi-objective optimization and data mining.
received the M.S. degree from Harbin Institute of Technology, Harbin, China, in 1990, and the Ph.D. degree from Harbin Engineering University, Harbin, China, in 2006. She is a professor and Ph.D. supervisor at the School of Information Engineering, Minzu University of China, Beijing, China. Her research interests include evolutionary computation, multi-objective optimization, data mining, and intelligent transportation systems.
\end{IEEEbiography}
\begin{IEEEbiography}[{\includegraphics[width=2in,height=1.25in,clip,keepaspectratio]{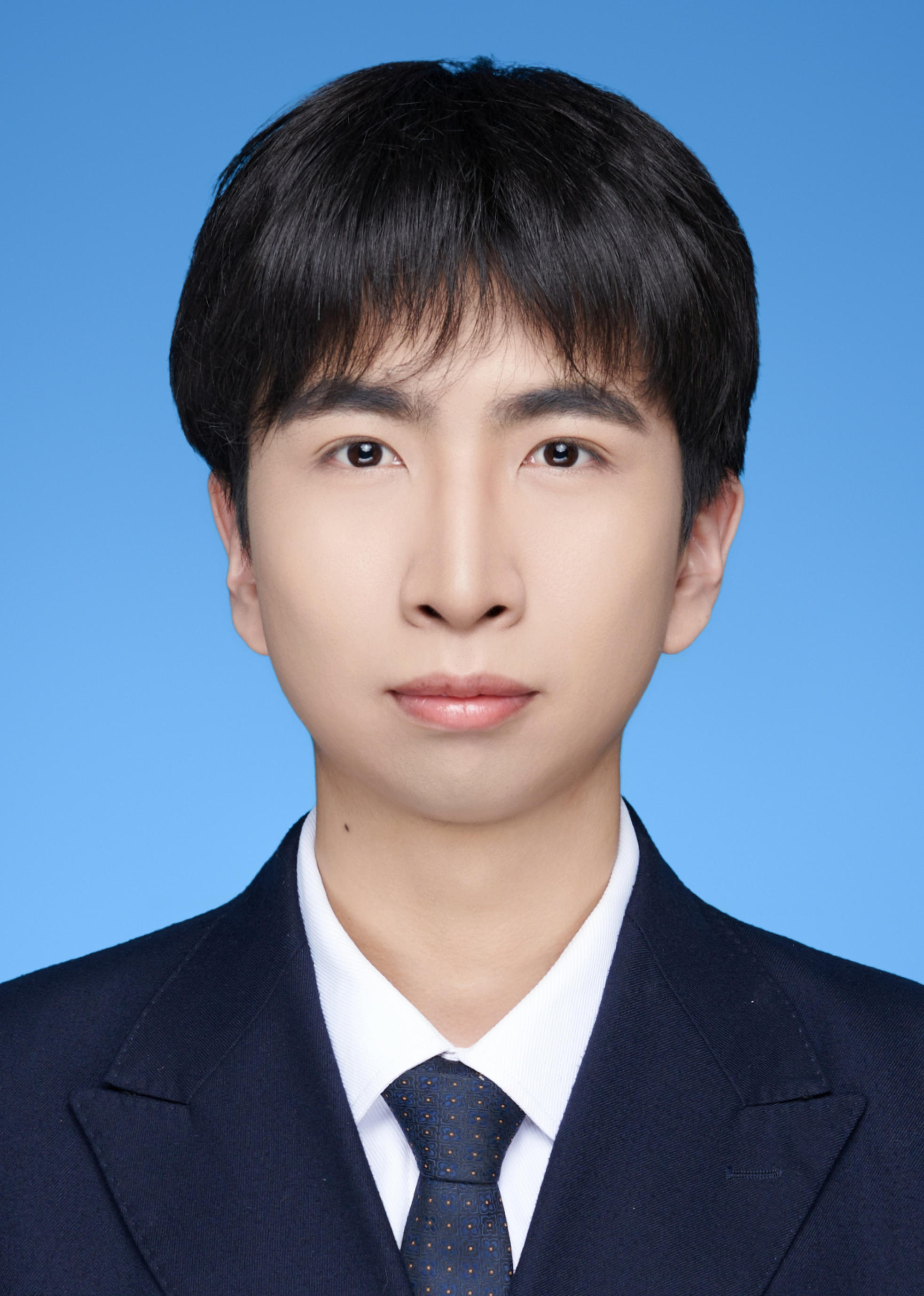}}]{Mingjie He}
received the B.Eng. Degree from Minzu University of China, China, in 2021. He is currently pursuing the M.Eng. degree in Minzu University of China. His research interests include deep reinforcement learning and intelligent transportation systems.\end{IEEEbiography}
\begin{IEEEbiography}[{\includegraphics[width=1in,height=1.25in,clip,keepaspectratio]{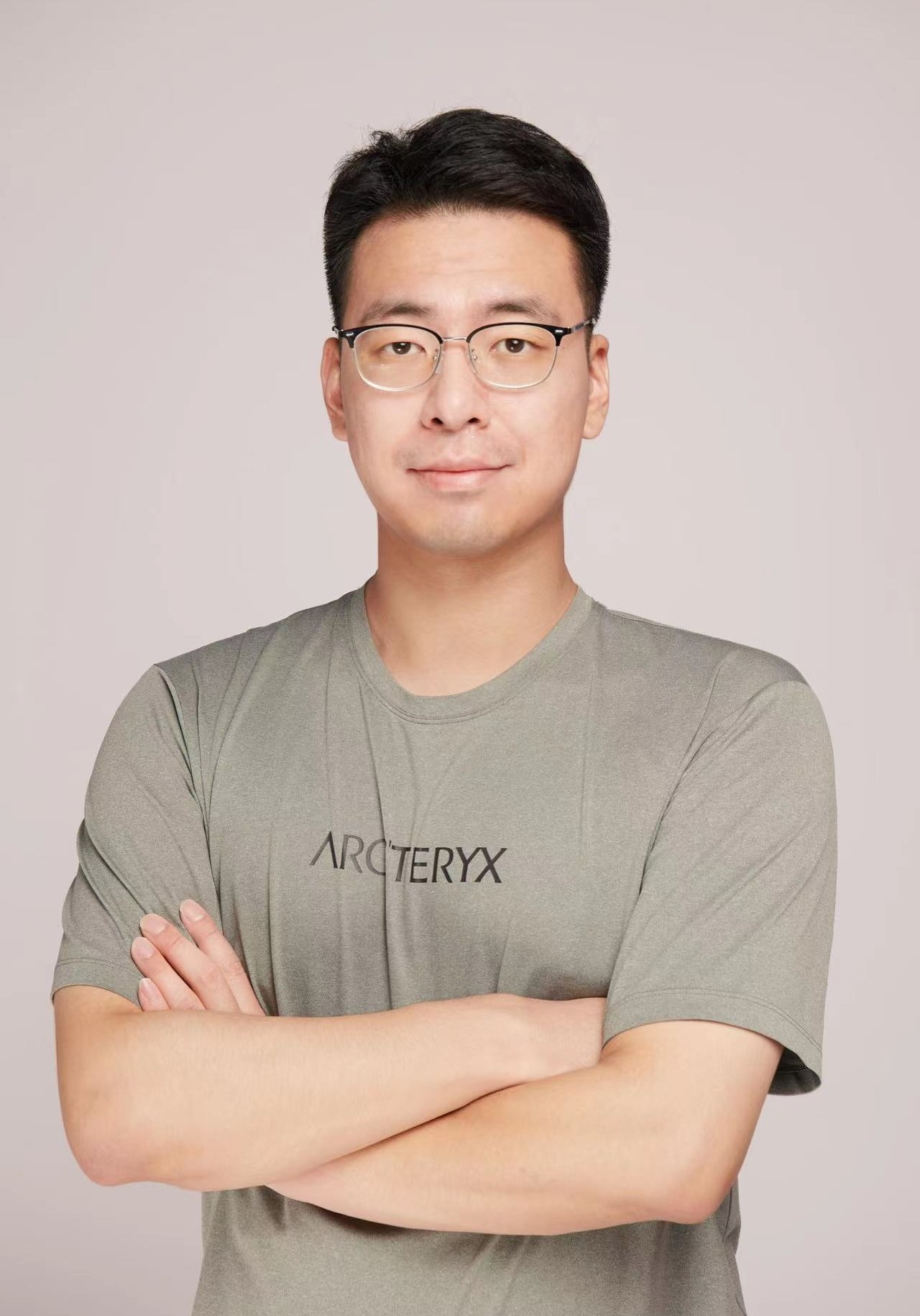}}]{Yiwen Sun}
received the Ph.D degree from the Department of Automation, Tsinghua University, Beijing, China, in 2022. He is currently a research assistant professor at the Institute for Artificial Intelligence, Peking University. His research interests include machine learning, sequence learning, spatial-temporal data mining, and intelligent transportation systems.\end{IEEEbiography}

\end{document}